\documentclass[lettersize,journal]{IEEEtran}
\usepackage{amsmath,amsfonts}
\usepackage{array}
\usepackage[caption=false,font=normalsize,labelfont=sf,textfont=sf]{subfig}
\usepackage{textcomp}
\usepackage{stfloats}
\usepackage{url}
\usepackage{verbatim}
\usepackage{graphicx}
\usepackage{cite}
\usepackage{color}
\usepackage{graphicx}
\usepackage{booktabs}

\usepackage{bm}
\usepackage{multirow}
\usepackage{makecell}
\usepackage{hyperref}
\usepackage[ruled,vlined]{algorithm2e}

\newcommand{\blue}[1] {\textcolor{black}{{#1}}}
\newcommand{\new}[1] {\textcolor{black}{{#1}}}

\begin{document}

\title{Forbid Your Attention: Fooling Multimodal Large Language Models by Selectively Removing Intrinsic Focus in Spectral Domain}

\author{Daizong Liu, Junhao Dong, Zhiyuan Ma, Xiaoye Qu, Xiang Fang, Runwei Guan, Keke Tang, \\ Jianfeng Dong, and Yew-Soon Ong,~\IEEEmembership{Fellow,~IEEE}
\IEEEcompsocitemizethanks{
\IEEEcompsocthanksitem This work was supported by the Fundamental and Interdisciplinary Disciplines Breakthrough Plan of the Ministry of Education of China (No. JYB2025XDXM101), the New Cornerstone Science Foundation through the XPLORER PRIZE, and the Innovative Research Group Project of Hubei Province under Grants 2024AFA017. This work was also supported by New Generation Artificial Intelligence-National Science and Technology Major Project (2025ZD0123602).
\IEEEcompsocthanksitem Daizong Liu is with the Institute for Math \& AI, Wuhan University, Wuhan, 430072, China. 
\protect
E-mail: daizongliu1996@gmail.com.
\IEEEcompsocthanksitem Junhao Dong, Xiang Fang, and Yew-Soon Ong are with the College of Computing and Data Science, Nanyang Technological University, Singapore. 
\protect
E-mail: junhao003@ntu.edu.sg, xfang9508@gmail.com, asysong@ntu.edu.sg.
\IEEEcompsocthanksitem Zhiyuan Ma and Xiaoye Qu are with the School of Computer Science and Technology, Huazhong University of Science and Technology, Wuhan, 430074, China.
\protect
E-mail: mzyth@hust.edu.cn, xiaoye@hust.edu.cn.
\IEEEcompsocthanksitem Runwei Guan is with the Department of Computer Science and Engineering, The Hong Kong University of Science and Technology, Guangzhou, 511455, China. 
\protect
E-mail: runwayrwguan@hkust-gz.edu.cn.
\IEEEcompsocthanksitem Keke Tang is with the Cyberspace Institute of Advanced Technology, Guangzhou University, Guangzhou, Guangdong, 510006, China. 
E-mail: tangbohutbh@gmail.com.
\IEEEcompsocthanksitem Jianfeng Dong is with the College of Computer and Information Engineering, Zhejiang Gongshang University, China. 
\protect
E-mail: dongjf24@gmail.com.
}
\thanks{Corresponding author: Xiang Fang.}}



\maketitle

\begin{abstract}
Multimodal large language models (MLLMs) have extended the capability of large language models (LLMs) to process more contextual multimodal information, showing remarkable progress in diverse realistic multimodal applications.
Despite their strong perception and reasoning abilities, recent studies reveal that MLLMs remain highly vulnerable to adversarial inputs, especially those targeting visual components. However, existing attacks mainly focus on global perturbations, lacking an understanding of how MLLMs internally interpret visual structures.
In this paper, we make the attempt to investigate the intrinsic focus of MLLMs in the frequency domain and discover that their predictions are particularly sensitive to phase information, which encodes essential structural and semantic cues. Based on this observation, we propose a novel phase-aware adversarial attack framework that explicitly restricts adversarial perturbations to structure-relevant phase regions to suppress the MLLMs' focus for effective and imperceptible attacks. To further amplify the structural influence, we also introduce an auxiliary adversarial prompt learning module to guide multimodal misalignment around phase-sensitive regions, misleading the MLLM’s attention toward targeted structural patterns. Extensive experiments on multiple representative MLLM models and datasets demonstrate the superior effectiveness of our method compared to existing attacks.
\end{abstract}

\begin{IEEEkeywords}
Multimodal large language models, spectral domain, adversarial attack
\end{IEEEkeywords}

\section{Introduction}
\label{sec:intro}
\IEEEPARstart{M}{ultimodal} large language models (MLLMs)  \cite{li2023blip,liu2024visual} have demonstrated significant capabilities in understanding and reasoning over diverse inputs from both visual and textual modalities, enabling strong performance on tasks such as visual question answering (VQA) \cite{xu2026ltkt,li2023blip,alayrac2022flamingo}, image captioning \cite{ye2026predicting,liu2021context,liu2020jointly,liu2022memory,jiang2025novel,liu2021adaptive,liu2022unsupervised,liu2022reducing}, and multimodal dialogue \cite{wang2025giving,lv2026adaptive,rombach2022high,chen2024vision,qi2025vision}. By leveraging the success of large language models (LLMs) \cite{brown2020language,touvron2023llama,touvron2023llama2} in text generation and alignment, MLLMs achieve impressive semantic comprehension across modalities, making them increasingly prevalent in critical real-world applications. However, recent works \cite{liu2024survey,wang2025invisible,hu2026trajectory,wang2025exploring,liu2024pandora,yan2025fit,cai2025towards,liu2026spatial,liu2026large,liu2026generating} have shown that MLLMs, like their unimodal counterparts, are susceptible to adversarial inputs—raising serious concerns about their reliability in real-world scenarios.

\begin{figure}[t!]
    \centering
    \includegraphics[width=0.48\textwidth]{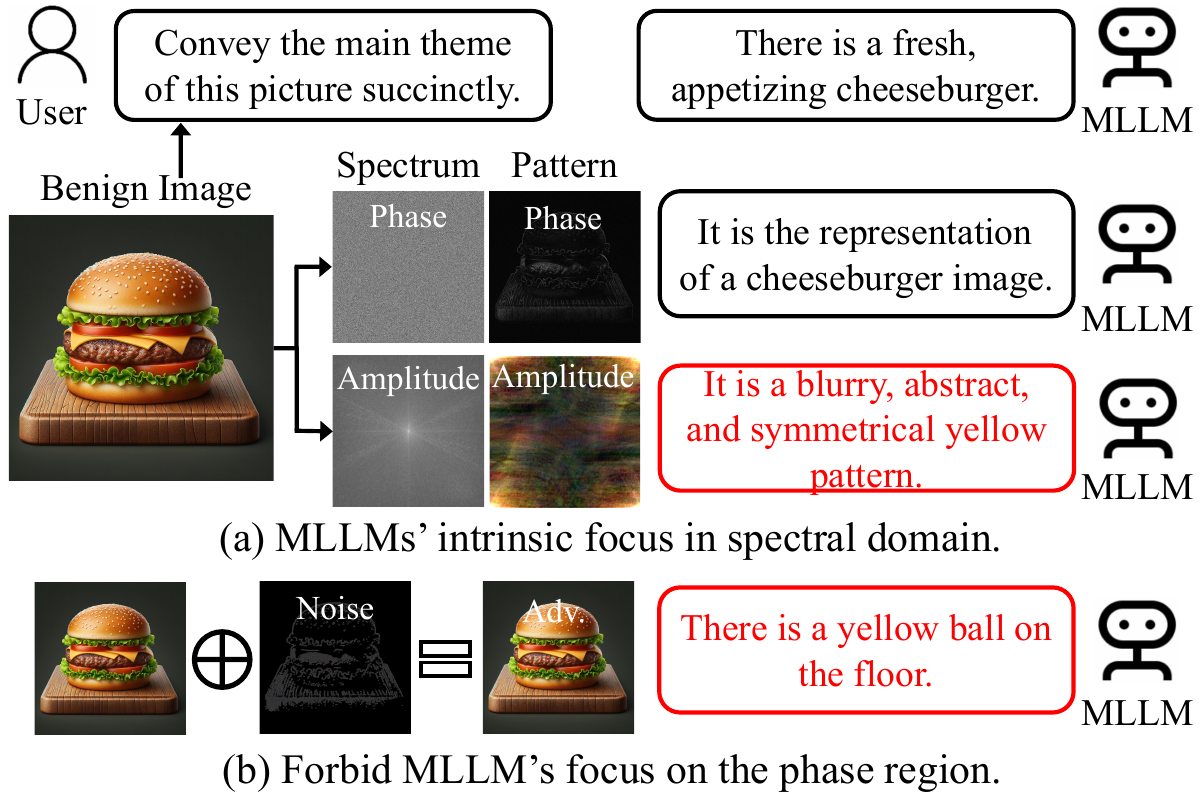}
    \caption{Illustration of our motivation. We empirically find that MLLMs rely more on the image's phase pattern for visual recognition and understanding. By suppressing MLLMs' corresponding focus, we can easily mislead their reasoning results.}
    \label{fig:intro}
\end{figure}

Prior studies \cite{shayegani2023jailbreak,bailey2023image,dong2023robust,wang2023instructta,wang2024stop,zhang2024avibench,lu2024test,luo2024image,tao2024imgtrojan,zhao2024evaluating} on adversarial robustness of MLLMs primarily focus on designing visual perturbations that implicitly interfere with the model's cross-modal reasoning process. Most of them \cite{bailey2023image,wang2024stop,lu2024test,luo2024image,tao2024imgtrojan} typically apply global pixel-level noise to the entire image, aiming to confuse the model’s multimodal alignment. 
Some works \cite{shayegani2023jailbreak,dong2023robust,wang2023instructta,zhao2024evaluating,zhang2024avibench} attempt to directly
mimic the semantic-irrelevant features by aligning the representation with pre-defined adversarial objectives, requiring access to the visual/textual encoder of MLLMs to learn the perturbation direction for misleading the latter's reasoning process. 
Despite this progress, existing MLLM attacks often overlook the intrinsic visual representations that MLLMs rely on internally for semantic grounding. Most approaches treat the visual input as a raw signal for gradient-based manipulation, without considering how different components of the visual information—such as spatial structures or frequency cues—contribute to the model's multimodal alignment. 
That is, systematically understanding the intrinsic focus of MLLMs—particularly the visual patterns they explicitly attend to during inference—remains an underexplored direction, which is essential for crafting more interpretable and effective adversarial strategies.

To design more principled adversarial attacks guided by how these MLLMs internally process visual information, we aim to investigate and exploit the intrinsic focus of MLLMs. Inspired by human visual perception \cite{oppenheim2005importance,karout2007two,guo2008spatio,li2015finding}, which identifies and interprets objects primarily through structural cues such as edges, contours, and spatial layouts, we hypothesize that MLLMs may exhibit a similar preference when grounding semantics in images. Specifically, decades of perceptual research \cite{oppenheim2005importance,karout2007two,guo2008spatio,li2015finding} have shown that the human visual system heavily relies on low-frequency phase information to recognize and reason about visual scenes, rather than the high-frequency amplitude details.
Following this intuition, we analyze images through the lens of their frequency components—decomposing them into amplitude and phase spectra via the Discrete Fourier Transform. As shown in Figure~\ref{fig:intro}, while the amplitude primarily captures stylistic and texture-related signals, the phase encodes geometric structures and spatial semantics \cite{chen2021amplitude}. To test whether MLLM models attend to these components differently, we construct controlled visual inputs that selectively modify either phase or amplitude while keeping the other fixed in Section~\ref{sec:emp}.
Our empirical observations reveal a striking finding: perturbations in the phase spectrum alone are sufficient to fool MLLMs, while amplitude perturbations result in significantly weaker or no adversarial effects. This suggests that MLLMs, like humans, implicitly rely on phase-encoded structural information to align visual content with textual queries. Such insights shed light on the internal mechanisms of MLLMs and offer a new direction for designing structure-aware and frequency-aligned adversarial attacks for MLLMs.

Based on the above observations, in this paper, we propose a novel phase-aware adversarial attack framework that exploits MLLMs' intrinsic reliance on phase-encoded structures. We first reconstruct a phase-only version of the input image and apply edge detection and adaptive thresholding to extract a binary structure mask that highlights phase-dominant regions. We then restrict the adversarial perturbation within these structural regions and optimize it using a multi-term loss function that combines task-specific adversarial loss, phase spectrum regularization function, and spatial-domain pattern divergence constraint. To further boost the effectiveness of the attack, we introduce an auxiliary adversarial prompt learning module that guides the MLLM’s attention toward the phase-sensitive regions, making the attack more targeted and semantically disruptive. Extensive experiments across multiple MLLMs demonstrate that our approach achieves significantly higher attack success rates than previous state-of-the-art methods.

In summary, our contributions are three-fold:
\begin{itemize}
\item We investigate MLLMs in the spectral domain and discover their intrinsic sensitivity to the phase spectrum, which governs semantic structure understanding.
\item We propose a novel phase-aware attack framework that applies perturbations to structurally meaningful regions and enforces spectrum-level variation, leading to more effective and interpretable attacks.
\item We design an auxiliary prompt optimization module with an adversarial learning strategy that cooperates with the phase perturbation to enhance the semantic disruption.
\end{itemize}

\section{Related Work}

\noindent \textbf{MLLM attacks.}
Existing multimodal large language models (MLLMs) attacks typically focus on implicitly disturbing the internal multimodal reasoning process by objective optimization \blue{\cite{qian2026individual,qian2025exploiting,qian2025multimodal,lai2025enhancing,fan2025underwater,zhu2026data}}.
Most studies~\cite{bailey2023image,wang2024stop,lu2024test,luo2024image,tao2024imgtrojan,liu2026attacking} propose to mislead MLLM predictions by applying pixel-level noise to the entire image, with the implicit constraint in the adversarial loss to push the response away from the accurate texts or pull the response close to an attacker-chosen target. Departing from these paradigms, the HardPatch~\cite{ai2026attacking} exploits the way MLLMs themselves partition and prioritize visual content, concentrating the whole perturbation budget on the few patches that dominate the model's perceptual focus, turning the perceptual prior of MLLMs from an implicit by-product of optimization into an explicit attack surface.
Another line of research \cite{shayegani2023jailbreak,dong2023robust,wang2023instructta,zhao2024evaluating,zhang2024avibench} attempts to mimic harmful semantics, hijack attention maps, or align features with pre-defined adversarial objectives, requiring access to the visual/textual encoder of MLLMs like CLIP \cite{radford2021learning,sun2023eva} to generate the perturbed visual representations for misleading the latter reasoning process. 
While these approaches have achieved notable adversarial effects, they tend to rely heavily on trial-and-error crafting or brute-force optimization, overlooking the intrinsic perceptual mechanisms of MLLMs and how MLLMS inherently perceive and organize visual inputs. Therefore, this paper investigates the MLLMs' intrinsic focus for adversarial designs.

\noindent \textbf{Frequency-based explanation.}
Frequency-domain analysis has been widely employed to explain and understand the behavior of deep neural networks \blue{\cite{qian2024enhancing,qian2025enhancing}}, particularly in the vision domain. Early studies revealed that convolutional neural networks (CNNs) tend to rely heavily on high-frequency components, making them vulnerable to imperceptible perturbations~\cite{yin2019fourier,wang2020high,liu2023point,hu2022exploring}. To better interpret model decisions, several works have proposed to decompose inputs into amplitude and phase spectra, showing that phase information encodes most of the semantic and structural content, whereas amplitude primarily captures style or texture~\cite{oppenheim2005importance,chen2021amplitude}. Based on this, amplitude-phase manipulations have been applied in model interpretation~\cite{luo2022frequency}, generalization analysis~\cite{geirhos2018imagenet}, and adversarial robustness studies~\cite{zhao2021understanding}.
In MLLM models, however, the frequency perspective remains underexplored. Most explanation efforts focus on cross-modal attention or token attribution, overlooking the frequency characteristics of the visual backbone that drive multimodal alignment. Our work is the first to empirically investigate MLLMs' sensitivity to frequency components, particularly phase, and leverage this to design a structure-aware attack.
\blue{Moreover, previous works simply decompose the frequency of input image into low, middle, and high bands to separately constrain the perturbations. However, we utilize amplitude-phase perspectives to disentangle the whole image into structure and styles. Further, our MLLM-based cases are more complicated than traditional classifiers, where both downstream tasks and language contexts provide additional reasoning challenges, such as VQA tasks requires special visual focus guided by the text. Therefore, our findings of MLLM’s intrinsic focus are specific.}

\begin{figure*}[t!]
    \centering
    \includegraphics[width=0.246\textwidth]{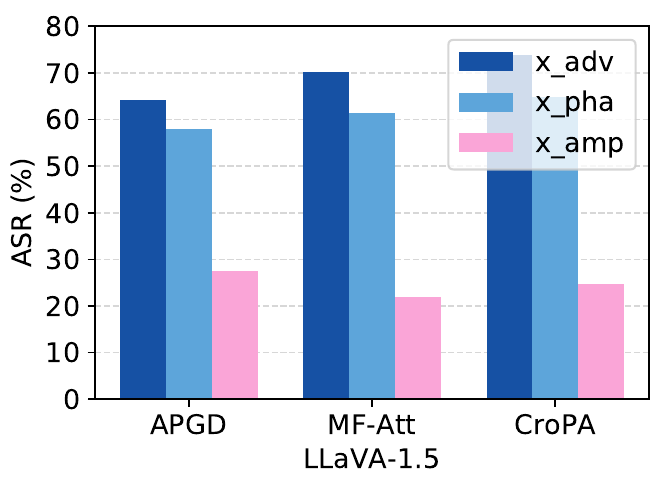}
    \hspace{-4pt}
    \includegraphics[width=0.246\textwidth]{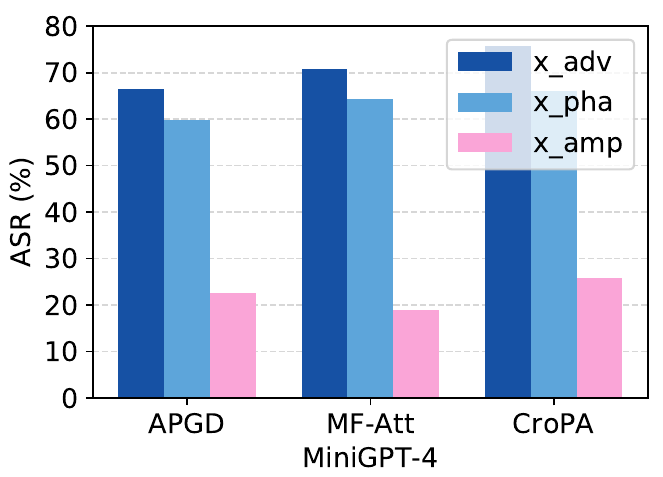}
    \hspace{-4pt}
    \includegraphics[width=0.246\textwidth]{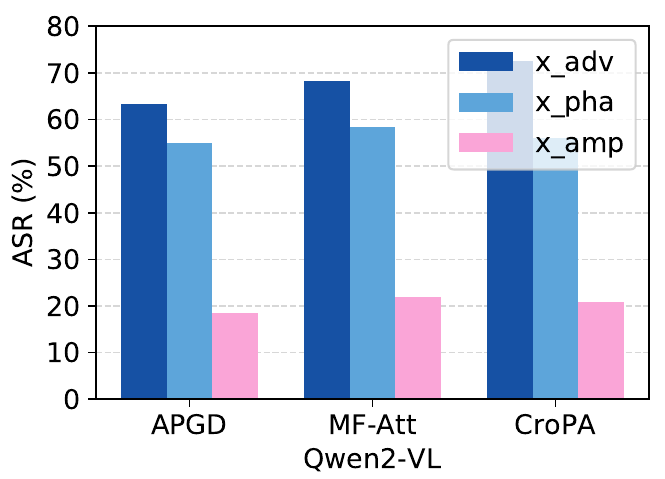}
    \hspace{-4pt}
    \includegraphics[width=0.246\textwidth]{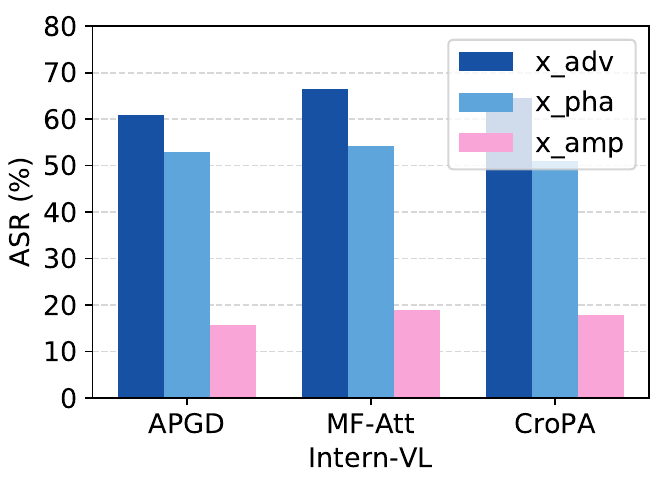}
    \vspace{-10pt}
    \caption{Attack success rate (ASR) of three MLLM attacks across four MLLM models. ``x\_adv" denotes the whole adversarial example, ``x\_pha" denotes phase-perturbed image $\bm{x}_{pha}'$, ``x\_amp" denotes amplitude-perturbed image $\bm{x}_{amp}'$.}
    \label{fig:empiric}
\end{figure*}

\begin{table*}[t]
\small
\centering
\caption{\blue{Controlled analysis of phase sensitivity in MLLMs.}}
\vspace{-6pt}
\begin{tabular}{c|cc|cc|cc}
\hline
\multirow{2}{*}{\blue{Model}}
&
\multicolumn{2}{c|}{\blue{Phase-Amplitude Swapping}}
&
\multicolumn{2}{c|}{\blue{Representation Similarity}}
&
\multicolumn{2}{c}{\blue{Attention Similarity}}
\\

&
\blue{Phase Follow}
&
\blue{Amplitude Follow}
&
\blue{Phase-only}
&
\blue{Amplitude-only}
&
\blue{Phase-preserved}
&
\blue{Amplitude-preserved}
\\
\hline

\blue{LLaVA-1.5}
& \blue{87.2} & \blue{12.8}
& \blue{0.87} & \blue{0.43}
& \blue{0.84} & \blue{0.44}
\\

\blue{MiniGPT-4}
& \blue{85.9} & \blue{14.1}
& \blue{0.85} & \blue{0.40}
& \blue{0.80} & \blue{0.42}
\\

\blue{Qwen2-VL}
& \blue{89.4} & \blue{10.6}
& \blue{0.89} & \blue{0.38}
& \blue{0.86} & \blue{0.41}
\\

\blue{Intern-VL}
& \blue{88.6} & \blue{11.4}
& \blue{0.88} & \blue{0.39}
& \blue{0.85} & \blue{0.43}
\\
\hline
\end{tabular}
\label{tab:phase_causal_analysis}
\end{table*}

\section{Investigation on MLLMs' Intrinsic Focus in Spectral Domain}
\label{sec:emp}
Multimodal large language models (MLLMs) generally receive a queried image $\bm{x}$ and a prompt $\bm{p}$ to generate corresponding response $\bm{r}$, \textit{i.e.}, $M(\bm{x},\bm{p})\rightarrow \bm{r}$. Due to the different types of prompt $\bm{p}$ (\textit{e.g.}, classification, captioning, or VQA), $\bm{p}$ can be either semantically relevant or irrelevant to the image $\bm{x}$. Therefore, to provide accurate responses $\bm{r}$, MLLMs are required to comprehensively comprehend the visual semantics for reasoning. 
Investigating such MLLMs' intrinsic focus on image inputs helps to develop robust and effective attack methods.

To understand an image, human visual perception \cite{oppenheim2005importance,karout2007two,guo2008spatio,li2015finding} operates through a hierarchical process, starting from the detection of low-level features such as edges, to the integration of high-level semantic concepts. 
Luckily, images' frequency spectra consisting of amplitude and phase provide corresponding contexts: the amplitude typically captures stylistic information, whereas the phase encompasses richer semantics \cite{chen2021amplitude}.
This frequency-based processing forms the basis for robust and efficient scene understanding.
Inspired by human perception, we argue that MLLMs also implicitly learn to attend to different frequency components across the image, guided by their alignment with textual semantics. This motivates a deeper investigation into their intrinsic focus in the spectral domain, which could reveal how these models internally prioritize visual information and how this spectral sensitivity correlates with their adversarial robustness or vulnerability.

\noindent \textbf{Sensitivity to phase/amplitude spectrum.}
Given an image $\bm{x}$, we utilize the Discrete Fourier Transform (DFT) $\mathcal{F}(\cdot)$ and inverse DFT (IDFT) functions $\mathcal{F}^{-1}(\cdot,\cdot)$ to obtain its phase/amplitude spectrum. Typically, DFT is independently applied to each channel of an image within the pixel space as:
\begin{equation}
    \mathcal{F}(\bm{x})(u,v) = \sum_{h=1}^H \sum_{w=1}^W \bm{x}(h,w)e^{-i2\pi(u\frac{h}{H}+v\frac{w}{W})},
\end{equation}
where $(h,w)$ denotes the pixel coordinates of $\bm{x}$, $(u,v)\in [H] \times [W]$ signifies coordinates in the frequency domain. Then, the amplitude spectrum $\mathcal{A}(\bm{x})$ and phase spectrum $\mathcal{P}(\bm{x})$ are obtained by:
\begin{equation}
\begin{aligned}
    &\mathcal{A}(\bm{x}) = (\text{Re}^2(\mathcal{F}(\bm{x}))+\text{Im}^2(\mathcal{F}(\bm{x})))^{\frac{1}{2}}, \\ & \mathcal{P}(\bm{x}) = \text{arctan}\frac{\text{Im}(\mathcal{F}(\bm{x}))}{\text{Re}(\mathcal{F}(\bm{x}))},
\end{aligned}
\end{equation}
where Re$(\cdot)$ and Im$(\cdot)$ denotes the real and imaginary parts.

To evaluate the sensitivity of MLLM models to the phase/amplitude spectrum, we first generate adversarial examples $\bm{x}'$ by three representative MLLM attacks, APGD \cite{cui2024robustness}, MF-Att \cite{zhao2024evaluating}, CroPA \cite{luo2024image}. Then, we utilize DFT to derive their phase and amplitude of the frequency spectra of both $\bm{x}$ and $\bm{x}'$. 
We compose phase- and amplitude-perturbed images $\bm{x}_{pha}',\bm{x}_{amp}'$ by combining adversarial phase/amplitude with benign amplitude/phase via:
\begin{equation}
  \bm{x}_{pha}'=\mathcal{F}^{-1}(\mathcal{A}(\bm{x}),\mathcal{P}(\bm{x}')), \bm{x}_{amp}'=\mathcal{F}^{-1}(\mathcal{A}(\bm{x}'),\mathcal{P}(\bm{x})).
\end{equation}
As shown in Figure~\ref{fig:empiric}, we can conclude that:
(1) Under all three attacks, the four MLLM models can be effectively fooled by the generated adversarial examples.
(2) $\bm{x}_{pha}'$ achieves a similar performance to the whole adversarial examples, while $\bm{x}_{amp}'$ achieves much worse performance.
These results suggest that MLLM models are more sensitive to the phase patterns than the amplitude ones.
The phenomena indicate that by forcing the model to focus less on the phase patterns of samples, the attack performance can be improved further.
\blue{To further investigate whether the observed attack effectiveness indeed originates from the intrinsic phase sensitivity of MLLMs, we conduct controlled analyses from three complementary perspectives. First, we perform phase-amplitude swapping experiments, where phase and amplitude spectra from two different images are recombined. We measure whether the model prediction follows the semantic content provided by the phase donor or the amplitude donor.
Second, we compare visual representations extracted from
MLLM encoders using phase-only and amplitude-only
reconstructions. Third, we analyze cross-modal attention similarity under phase-preserved and amplitude-preserved settings. Results are summarized in Table~\ref{tab:phase_causal_analysis}, suggesting that phase information plays a dominant role in semantic grounding and multimodal reasoning, providing stronger evidence beyond attack performance alone.}

\noindent \textbf{Discussion and motivation.}
Following the insight from the above studies, we propose to attack MLLM models by selectively removing their intrinsic focus in the spectral domain. Since the phase patterns generally reflect the highly informative structural features, we introduce to design phase-aware structure perturbations to explicitly suppress the MLLMs' focus on the phase spectrum.

\section{The Proposed Attack}

\subsection{Overview}
Given a MLLM model $M$ with image input $\bm{x}$ and prompt input $\bm{p}$,
our objective is to find an imperceptible visual perturbation $\bm{\delta}$ within a phase-aware structure region $\bm{m}$ such that the perturbed input $(\bm{x}+\bm{m}\odot \bm{\delta}, \bm{p})$ induces the MLLM model to generate a wrong response $\bm{r}', \bm{r}' \neq \bm{r}$.
$\bm{m}$ is generally the phase-guided image structure region that determines the MLLMs' comprehension.
We formulate this adversarial generation as a constrained optimization problem:
\begin{equation}\label{eq:main}
\min_{\bm{\delta}} \mathbb{E}_{(\bm{x},\bm{p})\sim\mathcal{D}} \left[ \mathcal{L}(M(\bm{x}+\bm{m}\odot \bm{\delta},\bm{p}), \bm{r}') \right], \|\bm{\delta}\|_\infty \leq \epsilon.
\end{equation}
Note that, we do not utilize $(1-\bm{m})\odot\bm{x}+\bm{m}\odot \bm{\delta}$ because it will introduce noticeable noise.

\begin{figure*}[t!]
    \centering
    \includegraphics[width=0.94\textwidth]{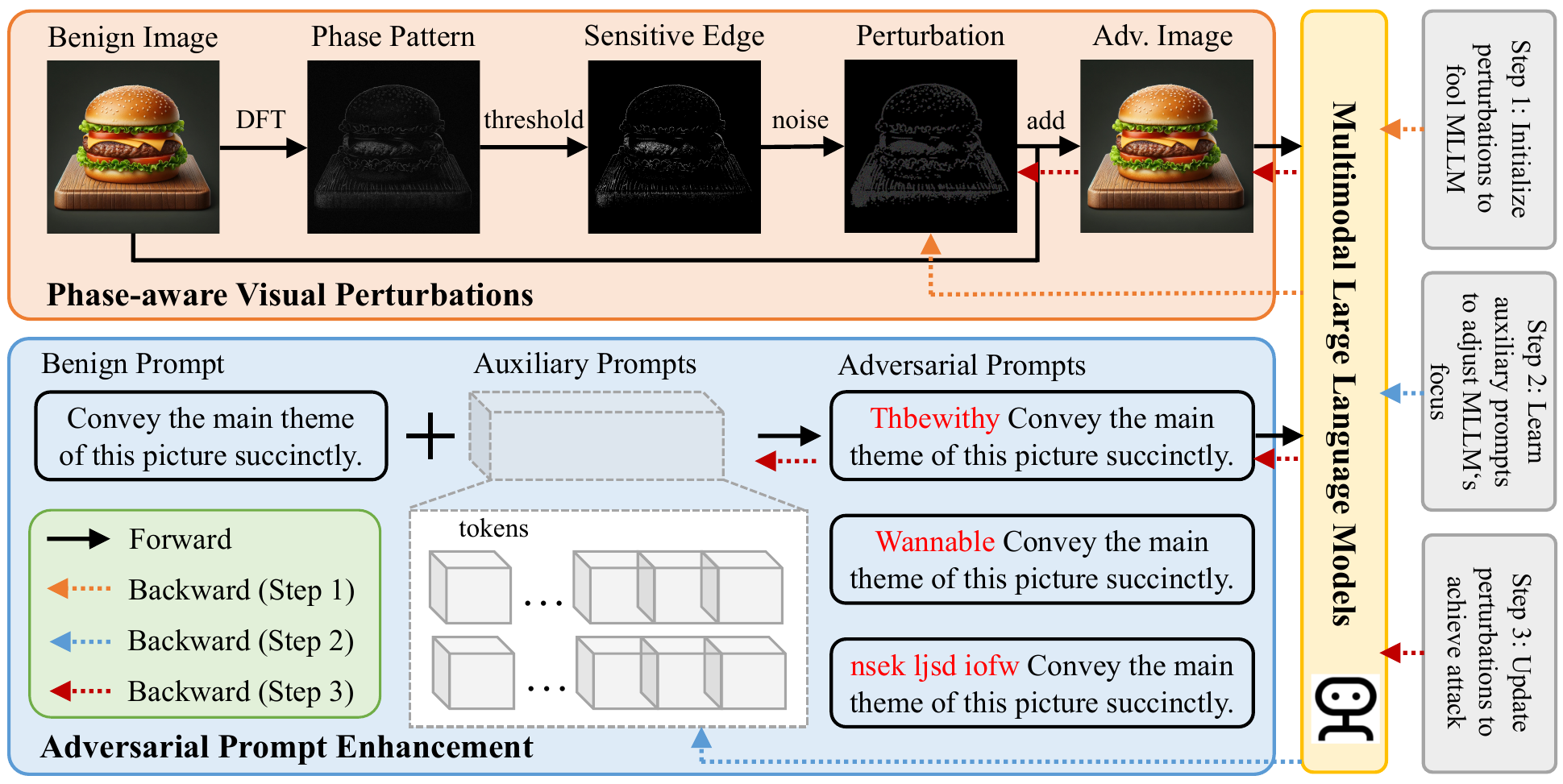}
    \caption{Overview of the proposed attack pipeline. We first generate MLLM's sensitive phase-aware perturbations to fool MLLM's intrinsic focus. Then, we further learn trainable auxiliary prompts to enhance the effectiveness of the phase-aware perturbations.}
    \label{fig:pipeline}
\end{figure*}

\noindent \textbf{Overall pipeline.}
To effectively mislead the MLLM's understanding and reasoning, we develop a novel attack method by suppressing the MLLM's intrinsic focus during the processing. The overall pipeline is illustrated in Figure~\ref{fig:pipeline}. We first utilize the DFT tool to obtain MLLM-sensitive phase-aware regions for perturbing the benign images. Although we can directly optimize these perturbations to fool the MLLM model, their impact on multimodal models may be diluted if the model’s attention is not sufficiently focused on these phase-dominant features. Therefore, we further develop trainable auxiliary prompts to enhance the harmfulness of the visual perturbations via an adversarial learning scheme. That is, we learn the adversarial prompts to guide MLLM models to focus less on the phase-aware perturbations, and in turn update the perturbations to bypass these prompts to achieve robust attacks.

\subsection{Perceiving Phase-aware Structure}
To effectively suppress MLLMs' sensitivity to phase components, we first extract the structural priors encoded within the phase spectrum. As demonstrated in prior works \cite{oppenheim2005importance,chen2021amplitude}, phase information preserves the majority of an image's geometric and semantic structure, such as object edges, contours, and spatial layouts.

Given an image $\bm{x}$, we perform DFT to extract its phase spectrum $\mathcal{P}(\bm{x})$. To highlight its structural content in the spatial domain, we reconstruct a phase-only image by discarding amplitude information and applying inverse DFT as:
\begin{equation}
\bm{x}_{pha} = \mathcal{F}^{-1}(\bm{1}, \mathcal{P}(\bm{x})),
\end{equation}
where $\bm{1}$ denotes an all-one amplitude matrix of the same shape as $\mathcal{A}(\bm{x})$. This operation ensures that the reconstruction relies solely on phase features, highlighting spatial discontinuities such as edges and contours.
We then convert the reconstructed image $\bm{x}_{pha}$ into a binary structure mask $\bm{m}$ that reflects salient phase-aware regions. This is achieved via gradient-based edge detection (\textit{e.g.}, Sobel filter) followed by adaptive thresholding:
\begin{equation}
\bm{m} = \text{Thresh}(\text{Sobel}(\bm{x}_{pha})),
\end{equation}
where $\text{Sobel}(\cdot)$ computes the spatial gradient magnitude, and $\text{Thresh}(\cdot)$ applies an adaptive binarization to isolate high-gradient (structural) areas.

The resulting mask $\bm{m}$ highlights the phase-dominant structural regions in $\bm{x}$, such as object contours and semantic boundaries, and will be used to guide perturbation localization in the following attack module. By focusing the adversarial perturbation within $\bm{m}$, we aim to maximally disrupt the visual cues that MLLMs rely on for semantic alignment and reasoning.

\subsection{Optimizing Phase-aware Perturbation}
Given the extracted phase-aware structure mask $\bm{m}$, our optimization goal is to generate an adversarial example $\bm{x}'= \bm{x} + \bm{\delta} \odot \bm{m}$ that preserves the perceptual similarity to the original image $\bm{x}$ while successfully misleading the target MLLM $M$ with respect to the input prompt $\bm{p}$. The perturbation is explicitly applied to structurally salient regions, as guided by $\bm{m}$.

\noindent \textbf{Task-oriented adversarial loss.}
To fool the MLLM models with downstream tasks, we define the attacker’s adversarial objective following Eq.(\ref{eq:main}) as:
\begin{equation}
\mathcal{L}_{\text{task}}(\bm{x}', \bm{p}, \bm{r}') = \mathcal{L}(M(\bm{x}', \bm{p}), \bm{r}'),
\end{equation}
where the task can be Visual Question Answering (VQA), Image Captioning, and Image Classification, \textit{etc.}. This term ensures that $\bm{x}'$ induces incorrect but plausible responses from the MLLM model $M$.

\noindent \textbf{Phase-domain spectrum regularization.}
To explicitly enforce structural changes in the frequency domain, we propose a phase variation loss that measures the deviation between the original and perturbed phase spectra in the masked region. Specifically, we define:
\begin{equation}
\mathcal{L}_{\text{phase}}(\bm{x}',\bm{x})  = \left| \bm{m} \odot \left( \mathcal{P}(\bm{x}') - \mathcal{P}(\bm{x}) \right) \right|_2^2,
\end{equation}
where $\mathcal{P}(\cdot)$ denotes the pixel-wise phase spectrum. This term encourages perturbations to induce meaningful structural changes within the phase-aware regions, which are known to dominate semantic understanding.

\noindent \textbf{Spatial-domain phase pattern regularization.}
To effectively mislead the MLLMs' focus on the phase spectrum, we further propose to optimize the phase pattern of the adversarial example to be far away from the benign one, suppressing the MLLMs reason with the accurate structural semantics:
\begin{equation}
\mathcal{L}_{\text{pattern}}(\bm{x}',\bm{x})  = -\left|  \mathcal{F}^{-1}(\bm{1}, \mathcal{P}(\bm{x}'))- \mathcal{F}^{-1}(\bm{1}, \mathcal{P}(\bm{x})) \right|_2^2.
\end{equation}

By jointly optimizing for adversarial effectiveness and phase-domain variation via:
\begin{equation}
    \bm{\delta} \leftarrow \text{clip}(\bm{\delta} \blue{-} \alpha\cdot \text{sign}(\nabla_{\bm{\delta}}(\mathcal{L}_{task}+\mathcal{L}_{phase}+\mathcal{L}_{pattern}))),
\end{equation}
the perturbation becomes more aligned with MLLMs’ intrinsic reliance on structural semantics, increasing the success rate of attacks.

\subsection{Adversarial Prompt Enhancement}
While phase perturbations inherently target structural semantics, their impact on multimodal models may be diluted if the model’s attention is not sufficiently focused on these phase-dominant features. To address this, we propose an adversarial prompt learning module, which serves as an auxiliary mechanism to enhance the optimization of visual phase perturbations.
Specifically, we introduce a trainable textual prompt $\bm{p}_{aux}$ that is designed to highlight phase-sensitive structures during the image-prompt interaction process. This auxiliary prompt is not adversarial by itself, but is jointly optimized to amplify the MLLM’s reliance on the masked structural regions $\bm{m}$ extracted from the image phase.

\begin{algorithm}[t!]
\caption{Our attack algorithm}
\label{alg:phase_attack}
\KwIn{Image $\bm{x}$, prompt $\bm{p}$, MLLM $M$, target response $\bm{r}'$, step size $\alpha$, numbers of iteration $T_1,T_2,T_3$}
\KwOut{Adversarial example $\bm{x}' = \bm{x} + \bm{\delta} \odot \bm{m}$}

Compute phase pattern of image $\bm{x}$: $\bm{x}_{pha} = \mathcal{F}^{-1}(\bm{1}, \mathcal{P}(\bm{x}))$ \\
Compute phase-aware structural mask: $\bm{m} = \text{Thresh}(\text{Sobel}(\bm{x}_{pha}))$ \\

\textbf{Initialize:} \\
$\bm{\delta} \leftarrow \bm{0}$\blue{, auxiliary prompt $\bm{p}_{aux}$ via random vocabulary high-frequency words}

\For{$t = 1$ \KwTo $T_1$}{
    adversarial example $\bm{x}' \leftarrow \bm{x} + \bm{\delta} \odot \bm{m}$ \\
    \For{$t = 1$ \KwTo $T_2$}{
    \blue{$\bm{p}_{aux} \leftarrow \text{Adam}\left(\bm{p}_{aux}, \nabla_{\bm{p}_{aux}} \mathcal{L}_{\text{task}}(\bm{x}', \bm{p} \oplus \bm{p}_{aux}, \bm{r})\right)$}
    }
    \For{$t = 1$ \KwTo $T_3$}{
    \blue{Compute $\mathcal{L}_{\text{total}} = \mathcal{L}_{\text{task}}(\bm{x}', \bm{p} \oplus \bm{p}_{aux}, \bm{r}') + \mathcal{L}_{\text{phase}} + \mathcal{L}_{\text{pattern}}$} \\
    $\bm{\delta} \leftarrow \text{clip}\left(\bm{\delta} \blue{-} \alpha \cdot \text{sign}(\nabla_{\bm{\delta}} \mathcal{L}_{\text{total}})\right)$
    }
}

\Return $\bm{x}' = \bm{x} + \bm{\delta} \odot \bm{m}$

\end{algorithm}

Concretely, given a benign image $\bm{x}$ and a structure mask $\bm{m}$, we first define a distortion loss that encourages the MLLM to ground its semantic understanding not on the masked phase regions when paired with $\bm{p}_{aux}$ to provide accurate response $\bm{r}$. Based on this, we optimize the $\bm{p}_{aux}$ via:
\begin{equation}
\mathcal{L}_{\text{task}}(\bm{x}', \bm{p}', \bm{r}) = \mathcal{L}(M(\bm{x}', \bm{p}'), \bm{r}),
\end{equation}
where $\bm{p}'=\bm{p} \oplus \bm{p}_{aux}$. We follow previous works \cite{zhou2022learning,li2024one} to define the learnable $\bm{p}_{aux}$ in the word token space. Then, during the optimization of the adversarial image $\bm{x}'$, we freeze the auxiliary prompt $\bm{p}_{aux}$ and define the adversarial loss to attack MLLMs with auxiliary prompt as:
\begin{equation}
\mathcal{L}=\mathcal{L}_{task}(\bm{x}',\bm{p}',\bm{r}')+\mathcal{L}_{phase}(\bm{x}',\bm{x})+\mathcal{L}_{pattern}(\bm{x}',\bm{x}).
\end{equation}

\begin{table*}[t!]
    \centering
    \caption{Targeted attack performance of different MLLM attack methods across multiple MLLM datasets and open-source models.}
    \vspace{-6pt}
    \resizebox{1.0\textwidth}{!}{
    \setlength{\tabcolsep}{0.6mm}{
    \begin{tabular}{c|c|ccc|ccc|ccc|ccc|ccc|ccc}
    \hline
    \multirow{2}*{Data} & \multirow{2}*{Attack} & \multicolumn{3}{c|}{BLIP-2 \cite{li2023blip}} & \multicolumn{3}{c|}{LLaVA-1.5 \cite{liu2024visual}} & \multicolumn{3}{c|}{Flamingo \cite{alayrac2022flamingo}} & \multicolumn{3}{c|}{MiniGPT-4 \cite{zhu2023minigpt}} & \multicolumn{3}{c|}{Qwen2-VL \cite{wang2024qwen2}} & \multicolumn{3}{c}{Intern-VL \cite{chen2024internvl}} \\
    ~ & ~ & SS & E-ASR & C-ASR & SS & E-ASR & C-ASR & SS & E-ASR & C-ASR & SS & E-ASR & C-ASR & SS & E-ASR & C-ASR & SS & E-ASR & C-ASR \\ \hline
     \multirow{6}*{\rotatebox{90}{ImageNet \cite{deng2009imagenet}}} & APGD \cite{cui2024robustness} & 0.662 & 61.0 & 66.2 & 0.670 & 60.2 & 65.0 & 0.655 & 59.1 & 63.4 & 0.685 & 63.5 & 67.0 & 0.676 & 61.8 & 65.2 & 0.661 & 60.5 & 64.1 \\
     ~ & MF-Att \cite{zhao2024evaluating} & 0.692 & 65.3 & 69.0 & 0.701 & 67.2 & 70.8 & 0.685 & 64.5 & 67.5 & 0.720 & 68.1 & 71.3 & 0.707 & 66.4 & 70.0 & 0.698 & 65.7 & 68.2 \\
     ~ & CroPA \cite{luo2024image} & 0.715 & 67.9 & 71.2 & 0.726 & 69.5 & 72.4 & 0.701 & 66.1 & 69.7 & 0.739 & 70.2 & 73.4 & 0.723 & 68.6 & 71.2 & 0.718 & 67.8 & 70.6 \\
     ~ & MABA \cite{liang2024revisiting} & 0.748 & 71.5 & 74.8 & 0.779 & 75.3 & 79.8 & 0.742 & 72.5 & 75.0 & 0.784 & 77.4 & 80.5 & 0.771 & 76.2 & 78.9 & 0.765 & 74.8 & 77.0 \\
     ~ & VMA \cite{wang2025attention} & 0.768 & 74.2 & 77.1 & 0.796 & 78.5 & 81.1 & 0.765 & 75.3 & 78.0 & 0.803 & 79.4 & 81.9 & 0.789 & 77.5 & 80.1 & 0.783 & 76.2 & 79.0 \\
     ~ & Ours & \textbf{0.841} & \textbf{86.5} & \textbf{88.2} & \textbf{0.904} & \textbf{90.0} & \textbf{90.9} & \textbf{0.855} & \textbf{86.2} & \textbf{87.1} & \textbf{0.899} & \textbf{88.7} & \textbf{89.4} & \textbf{0.895} & \textbf{88.3} & \textbf{88.7} & \textbf{0.887} & \textbf{87.6} & \textbf{88.5} \\ \hline
     \multirow{6}*{\rotatebox{90}{SVIT \cite{zhao2023svit}}} & APGD \cite{cui2024robustness} & 0.652 & 59.3 & 64.1 & 0.667 & 61.8 & 65.4 & 0.648 & 60.1 & 63.2 & 0.680 & 63.9 & 67.1 & 0.662 & 61.4 & 64.8 & 0.650 & 59.9 & 63.3\\
     ~ & MF-Att \cite{zhao2024evaluating} & 0.679 & 63.0 & 66.5 & 0.692 & 66.3 & 69.0 & 0.675 & 62.1 & 65.4 & 0.712 & 67.0 & 70.0 & 0.699 & 64.5 & 67.6 & 0.688 & 63.2 & 66.2\\
     ~ & CroPA \cite{luo2024image} & 0.703 & 66.8 & 69.5 & 0.720 & 68.2 & 71.4 & 0.693 & 65.4 & 67.9 & 0.732 & 69.1 & 71.7 & 0.715 & 67.8 & 70.5 & 0.709 & 66.4 & 68.8 \\
     ~ & MABA \cite{liang2024revisiting} & 0.738 & 70.4 & 73.5 & 0.765 & 74.4 & 77.2 & 0.725 & 70.6 & 73.0 & 0.788 & 76.9 & 78.5 & 0.765 & 74.4 & 77.2 & 0.757 & 73.5 & 75.4\\
     ~ & VMA \cite{wang2025attention} & 0.757 & 72.2 & 75.1 & 0.785 & 77.1 & 79.4 & 0.747 & 73.3 & 75.9 & 0.796 & 78.6 & 80.5 & 0.781 & 76.4 & 78.6 & 0.774 & 75.0 & 77.4 \\
     ~ & Ours & \textbf{0.832} & \textbf{85.0} & \textbf{86.8} & \textbf{0.893} & \textbf{89.5} & \textbf{90.7} & \textbf{0.848} & \textbf{85.1} & \textbf{85.9} & \textbf{0.901} & \textbf{89.3} & \textbf{90.2} & \textbf{0.889} & \textbf{87.9} & \textbf{89.3} & \textbf{0.882} & \textbf{86.5} & \textbf{88.1}\\ \hline
     \multirow{6}*{\rotatebox{90}{DALLE \cite{ramesh2021zero}}} & APGD \cite{cui2024robustness} & 0.701 & 66.9 & 72.5 & 0.701 & 66.9 & 72.5 & 0.695 & 65.3 & 70.4 & 0.725 & 68.2 & 74.9 & 0.734 & 67.8 & 72.7 & 0.715 & 66.3 & 71.1\\
     ~ & MF-Att \cite{zhao2024evaluating} & 0.728 & 70.3 & 74.1 & 0.735 & 71.2 & 75.5 & 0.719 & 69.0 & 72.8 & 0.752 & 73.0 & 76.2 & 0.741 & 71.4 & 74.8 & 0.728 & 69.5 & 72.6\\
     ~ & CroPA \cite{luo2024image} & 0.742 & 72.0 & 75.8 & 0.748 & 73.3 & 77.0 & 0.735 & 71.2 & 74.5 & 0.771 & 76.2 & 78.9 & 0.763 & 73.8 & 77.4 & 0.751 & 72.1 & 75.0\\
     ~ & MABA \cite{liang2024revisiting} & 0.779 & 75.3 & 79.8 & 0.779 & 75.3 & 79.8 & 0.767 & 73.1 & 77.0 & 0.784 & 77.4 & 80.5 & 0.771 & 76.2 & 78.9 & 0.765 & 74.8 & 77.0\\
     ~ & VMA \cite{wang2025attention} & 0.796 & 77.9 & 81.1 & 0.803 & 78.7 & 81.8 & 0.785 & 75.4 & 78.5 & 0.809 & 79.3 & 81.2 & 0.795 & 77.1 & 79.4 & 0.788 & 76.0 & 78.3\\
     ~ & Ours & \textbf{0.897} & \textbf{87.0} & \textbf{90.1} & \textbf{0.904} & \textbf{90.0} & \textbf{90.9} & \textbf{0.891} & \textbf{88.0} & \textbf{88.7} & \textbf{0.899} & \textbf{88.7} & \textbf{89.4} & \textbf{0.895} & \textbf{88.3} & \textbf{88.7} & \textbf{0.887} & \textbf{87.6} & \textbf{88.5} \\ \hline
    \end{tabular}}}
    \label{tab:main}
\end{table*}

The whole process is shown in Algorithm~\ref{alg:phase_attack}, where the auxiliary prompt acts as a phase-sensitive ``semantic lens", guiding the attack optimization to focus on perturbing the structural components that the model heavily relies on. 
\blue{The learnable prompt is only utilized to serve as an auxiliary tool during the attack optimization stage to enhance the effectiveness/robustness of the adversarial examples, which will not be utilized at inference.}
Unlike previous global attack methods, our MLLM-sensitive design preserves semantic plausibility, ensuring that the attack remains imperceptible from the language side while being more effective from the vision.

\subsection{\blue{Discussion on Novelty and Design Choice}}
\noindent\textbf{\blue{Relationship to prior frequency-domain attacks.}}
\blue{We acknowledge that some building blocks used in our framework, including Fourier analysis, structural extraction, and adversarial optimization, have been explored in previous adversarial attack literature. However, our work differs from existing frequency-domain attacks in both motivation and formulation.
Most prior frequency-based attacks are developed for conventional image classifiers and primarily focus on perturbing predefined frequency bands to improve imperceptibility, transferability, or robustness. In contrast, our work investigates the role of frequency information in multimodal semantic grounding and language reasoning. Rather than treating frequency as a perturbation constraint, we analyze how different spectral components contribute to MLLM perception and subsequently design attacks according to the observed semantic sensitivity.}

\noindent\textbf{\blue{Novelty of the proposed framework.}}
\blue{Previous studies have shown that phase information preserves structural content in natural images. However, whether such observations remain valid in multimodal large language models has not been systematically investigated. Our controlled analyses demonstrate that MLLMs exhibit substantially stronger dependence on phase information than amplitude information at the prediction, representation, and cross-modal attention levels. This observation motivates the proposed phase-aware semantic attack framework. The novelty of our method does not stem from any individual optimization component. Instead, it lies in establishing a phase-aware semantic grounding perspective for MLLM attacks and integrating frequency-domain analysis with multimodal semantic optimization. To the best of our knowledge, this is among the first studies that systematically connect spectral sensitivity, semantic grounding, and adversarial robustness in multimodal large language models.}

\section{Experiments}

\subsection{Experimental Setup}

\noindent \textbf{MLLM datasets.}
We utilize ImageNet \cite{deng2009imagenet}, SVIT \cite{zhao2023svit}, and DALLE \cite{ramesh2021zero} datasets for image captioning, image classification, and VQA tasks. 
\blue{As for evaluation metrics, we (1) measure the semantic similarity (SS) between raw and adversarial answers; 
(2) measure the ``ExactMatch" attack success rates  (E-ASR) to assess word-to-word overlap between output and target texts; (3) measure the ``Conditional-Contain" success rates (C-ASR)  to assess a flexible word-level overlap between output and target texts.}

\noindent \textbf{MLLM models.}
To assess the MLLMs' robustness against our attack,
we select six open-source MLLM models with different architectures for evaluation: BLIP-2 \cite{li2023blip}, LLaVA-1.5 \cite{liu2024visual}, Flamingo \cite{alayrac2022flamingo}, MiniGPT-4 \cite{zhu2023minigpt}, Qwen2-VL \cite{wang2024qwen2}, and Intern-VL \cite{chen2024internvl}. Five closed-source MLLM models, Claude-3.5 \cite{anthropic2024claude}, Claude-3.7 \cite{anthropic2025claude}, GPT-4o \cite{openai2024gpt4o}, GPT-4.1 \cite{openai2025gpt4}, Gemini-2.0 \cite{deepmind2024gemini2}, are also implemented.

\noindent \textbf{Implementation details.}
For attack setup, we utilize a maximum of $T_1=100$ epochs to optimize the adversarial noise and set the perturbation budget $\epsilon=8$, the decay factor $\mu=0.9$, the step size $\alpha=\epsilon/epoch$. To implement the adversarial learning, we set $T_2=30, T_3=20$ to optimize the perturbation. 
\blue{For each benchmark (ImageNet, SVIT, and DALLE), we randomly sample 1,000 images and use the same evaluation subset for all compared methods. Input images are resized to $224\times224$ before optimization.
For phase-aware perturbation generation, we first construct phase-dominant regions by reconstructing a phase-only image in the Fourier domain and extracting the corresponding structural response map. The top 20\% highest-response pixels are selected as phase-dominant regions. 
For targeted attacks, target labels or responses are randomly selected from semantically unrelated categories or a predefined target pool. The same target assignments are shared across all compared methods to ensure fair evaluation. For closed-source MLLMs, we use official APIs with deterministic decoding (temperature = 0) and a maximum query budget of 100 queries per image.
All compared baselines are implemented using their official releases whenever available and are evaluated under the same perturbation budget and experimental protocol as us for fair comparison. To assess statistical stability, each experiment is repeated using three random seeds \{0,1,2\}.}
All experiments are conducted on a single NVIDIA H100 Tensor Core GPU.

\begin{table*}[t!]
    \caption{Targeted attack performance of different MLLM attack methods across multiple MLLM datasets and closed-source models.}
    \vspace{-6pt}
    \resizebox{1.0\textwidth}{!}{
    \setlength{\tabcolsep}{1.2mm}{
    \begin{tabular}{c|c|ccc|ccc|ccc|ccc|ccc}
    \hline
    \multirow{2}*{Data} & \multirow{2}*{Attack} & \multicolumn{3}{c|}{Claude-3.5 \cite{anthropic2024claude}} & \multicolumn{3}{c|}{Claude-3.7 \cite{anthropic2025claude}} & \multicolumn{3}{c|}{GPT-4o \cite{openai2024gpt4o}} & \multicolumn{3}{c|}{GPT-4.1 \cite{openai2025gpt4}} & \multicolumn{3}{c}{Gemini-2.0 \cite{deepmind2024gemini2}} \\
    ~ & ~ & SS & E-ASR & C-ASR & SS & E-ASR & C-ASR & SS & E-ASR & C-ASR & SS & E-ASR & C-ASR & SS & E-ASR & C-ASR \\ \hline
     \multirow{5}*{\rotatebox{90}{ImageNet}} & MF-Att \cite{zhao2024evaluating} & 0.22 & 6.8 & 10.3 & 0.24 & 7.4 & 11.5 & 0.20 & 6.2 & 9.1 & 0.21 & 6.5 & 9.8 & 0.19 & 5.9 & 8.5 \\
     ~ & M-Att \cite{li2025frustratingly} & 0.39 & 28.3 & 32.4 & 0.41 & 30.2 & 34.1 & 0.38 & 27.5 & 30.7 & 0.37 & 25.8 & 29.5 & 0.36 & 24.3 & 27.1 \\
     ~ & \blue{M-Att v2 \cite{zhao2026pushing}} & \blue{0.44} & \blue{32.1} & \blue{35.2} & \blue{0.46} & \blue{34.0} & \blue{37.3} & \blue{0.47} & \blue{35.6} & \blue{40.8} & \blue{0.45} & \blue{33.7} & \blue{38.1} & \blue{0.42} & \blue{33.9} & \blue{36.5}\\
     ~ & FOA \cite{jia2025adversarial} & 0.47 & 34.8 & 40.2 & 0.49 & 38.5 & 43.0 & 0.46 & 35.2 & 39.6 & 0.44 & 33.0 & 38.2 & 0.43 & 31.6 & 36.0 \\
     ~ & Ours & \textbf{0.58} & \textbf{53.7} & \textbf{59.8} & \textbf{0.60} & \textbf{57.2} & \textbf{63.1} & \textbf{0.59} & \textbf{54.8} & \textbf{60.2} & \textbf{0.57} & \textbf{52.3} & \textbf{58.6} & \textbf{0.55} & \textbf{49.1} & \textbf{55.7} \\ \hline
     \multirow{5}*{\rotatebox{90}{SVIT}} & MF-Att \cite{zhao2024evaluating} & 0.24 & 7.5 & 11.2 & 0.25 & 8.2 & 12.4 & 0.22 & 7.2 & 10.6 & 0.23 & 7.7 & 11.3 & 0.21 & 6.8 & 9.7\\
     ~ & M-Att \cite{li2025frustratingly} & 0.41 & 30.3 & 35.1 & 0.43 & 33.1 & 37.4 & 0.40 & 29.0 & 33.6 & 0.39 & 27.8 & 32.5 & 0.38 & 26.1 & 30.1\\
     & \blue{M-Att v2 \cite{zhao2026pushing}} & \blue{0.46} & \blue{35.8} & \blue{40.5} & \blue{0.48} & \blue{38.2} & \blue{43.0} & \blue{0.47} & \blue{36.9} & \blue{41.6} & \blue{0.45} & \blue{35.1} & \blue{39.8} & \blue{0.43} & \blue{33.0} & \blue{37.4} \\
     ~ & FOA \cite{jia2025adversarial} & 0.49 & 38.6 & 44.0 & 0.51 & 41.2 & 47.1 & 0.48 & 38.0 & 43.3 & 0.46 & 36.0 & 41.2 & 0.45 & 34.2 & 39.0\\
     ~ & Ours & \textbf{0.61} & \textbf{58.1} & \textbf{64.4} & \textbf{0.63} & \textbf{61.2} & \textbf{66.8} & \textbf{0.61} & \textbf{57.5} & \textbf{63.3} & \textbf{0.60} & \textbf{55.6} & \textbf{61.5} & \textbf{0.58} & \textbf{52.1} & \textbf{58.7}\\ \hline
     \multirow{5}*{\rotatebox{90}{DALLE }} & MF-Att \cite{zhao2024evaluating} & 0.26 & 8.6 & 12.7 & 0.27 & 9.1 & 13.5 & 0.24 & 8.1 & 11.9 & 0.25 & 8.3 & 12.3 & 0.23 & 7.4 & 10.8\\
     ~ & M-Att \cite{li2025frustratingly} & 0.42 & 31.7 & 36.5 & 0.44 & 33.5 & 38.8 & 0.41 & 30.3 & 35.1 & 0.39 & 28.6 & 33.7 & 0.38 & 26.8 & 31.4\\
     ~ & \blue{M-Att v2 \cite{zhao2026pushing}} & \blue{0.48} & \blue{38.1} & \blue{43.7} & \blue{0.50} & \blue{40.6} & \blue{46.1} & \blue{0.49} & \blue{39.2} & \blue{45.0} & \blue{0.47} & \blue{37.5} & \blue{43.2} & \blue{0.45} & \blue{35.3} & \blue{40.5} \\
     ~ & FOA \cite{jia2025adversarial} & 0.50 & 40.9 & 46.8 & 0.52 & 43.3 & 48.9 & 0.49 & 39.2 & 45.5 & 0.47 & 37.6 & 43.8 & 0.45 & 35.1 & 41.0\\
     ~ & Ours & \textbf{0.63} & \textbf{61.3} & \textbf{67.9} & \textbf{0.65} & \textbf{64.7} & \textbf{70.2} & \textbf{0.63} & \textbf{60.1} & \textbf{67.0} & \textbf{0.61} & \textbf{57.2} & \textbf{64.5} & \textbf{0.60} & \textbf{54.3} & \textbf{61.8}\\ \hline
    \end{tabular}}}
    \label{tab:main2}
\end{table*}

\begin{table*}[t!]
    \centering
    \caption{Ablation study on each component of our proposed method on the ImageNet dataset \cite{deng2009imagenet}.}
    \vspace{-6pt}
    \resizebox{1.0\textwidth}{!}{
    \setlength{\tabcolsep}{0.8mm}{
    \begin{tabular}{ccccc|ccc|ccc|ccc|ccc}
    \hline
    Phase & \multicolumn{3}{c}{Optimization} & Adversarial & \multicolumn{3}{c|}{BLIP-2 \cite{li2023blip}} & \multicolumn{3}{c|}{LLaVA-1.5 \cite{liu2024visual}} & \multicolumn{3}{c|}{Qwen2-VL \cite{wang2024qwen2}} & \multicolumn{3}{c}{Intern-VL \cite{chen2024internvl}} \\ \cline{2-4}
    Region & $\mathcal{L}_{task}$ & $\mathcal{L}_{phase}$ & $\mathcal{L}_{pattern}$ & Learning & SS & E-ASR & C-ASR & SS & E-ASR & C-ASR & SS & E-ASR & C-ASR & SS & E-ASR & C-ASR  \\ \hline
    $\times$ & $\checkmark$ & $\times$ & $\times$ & $\times$ 
             & 0.541 & 47.2 & 50.8 
             & 0.592 & 54.5 & 56.1 
             & 0.574 & 51.0 & 52.7 
             & 0.563 & 49.6 & 51.2 \\
             
    $\checkmark$ & $\checkmark$ & $\times$ & $\times$ & $\times$ 
                & 0.765 & 73.6 & 76.4 
                & 0.825 & 81.1 & 82.5 
                & 0.814 & 78.2 & 79.3 
                & 0.802 & 77.5 & 78.4 \\

    $\checkmark$ & $\checkmark$ & $\checkmark$ & $\times$ & $\times$ 
                & 0.742 & 70.1 & 74.8 
                & 0.801 & 78.3 & 80.2 
                & 0.789 & 75.6 & 77.2 
                & 0.777 & 74.4 & 75.9 \\

    $\checkmark$ & $\checkmark$ & $\checkmark$ & $\checkmark$ & $\times$ 
                & 0.781 & 76.8 & 79.0 
                & 0.849 & 85.3 & 86.5 
                & 0.838 & 82.4 & 83.1 
                & 0.825 & 81.6 & 82.2 \\

    $\checkmark$ & $\checkmark$ & $\checkmark$ & $\checkmark$ & $\checkmark$ 
                & \textbf{0.841} & \textbf{86.5} & \textbf{88.2} 
                & \textbf{0.904} & \textbf{90.0} & \textbf{90.9} 
                & \textbf{0.895} & \textbf{88.3} & \textbf{88.7} 
                & \textbf{0.887} & \textbf{87.6} & \textbf{88.5} \\ \hline
    \end{tabular}}}
    \label{tab:ablation1}
\end{table*}

\subsection{Main Attack Performance}
As shown in Table~\ref{tab:main}, we evaluate the proposed method and five competitive baselines (APGD \cite{cui2024robustness}, MF-Att \cite{zhao2024evaluating}, CroPA \cite{luo2024image}, MABA \cite{liang2024revisiting}, VMA \cite{wang2025attention}) on three public datasets, targeting six popular open-source MLLM models. 
\blue{During adversarial example generation, the attacker has access to gradients, logits, probabilities, internal activations, visual encoder parameters, and tokenizer information from the target model.}
Across all configurations, our method consistently achieves the highest attack success rates on all evaluation metrics. This demonstrates the capability of our method to generalize well across diverse architectures and datasets.

To further assess the effectiveness of our method under more practical and challenging conditions, \blue{we extend our method with the Monte Carlo method \cite{james1980monte} to handle the black-box scenarios.
It can only access final textual responses through official APIs. No model internals, confidence scores, token probabilities, or gradient information are accessible.}
We evaluate the targeted attack performance on five widely used MLLM models across three datasets.
Since baselines in Table~\ref{tab:main} cannot handle the black-box MLLMs, we implement \blue{five} black-box baselines for comparison.
As reported in Table~\ref{tab:main2}, our method also significantly outperforms state-of-the-art \blue{black-box attack} baselines. These results confirm the high attack generality of our approach, making it highly applicable to real-world scenarios.

\subsection{Ablation Study}

\noindent \textbf{Effectiveness of each component.} 
We conduct an ablation study to assess the effectiveness of each component in our proposed framework. As shown in Table~\ref{tab:ablation1}, starting from a baseline with only the task loss, we incrementally add different modules to examine their individual impact. The phase region module leads to the most notable improvement, indicating that constraining perturbations to perceptually and semantically important frequency regions is crucial for enhancing attack success.
Further incorporating the pattern consistency loss improves the alignment between the generated perturbation and the desired structural semantics, leading to more coherent adversarial responses. Adversarial learning contributes additional robustness by optimizing perturbations under a range of simulated conditions, resulting in stronger transferability. Applying a phase regularization loss slightly hampers performance, this is because its constraint on perturbation magnitude, which limits the expressiveness of the attack. 

\begin{figure*}[t!]
    \centering
    \includegraphics[width=0.246\textwidth]{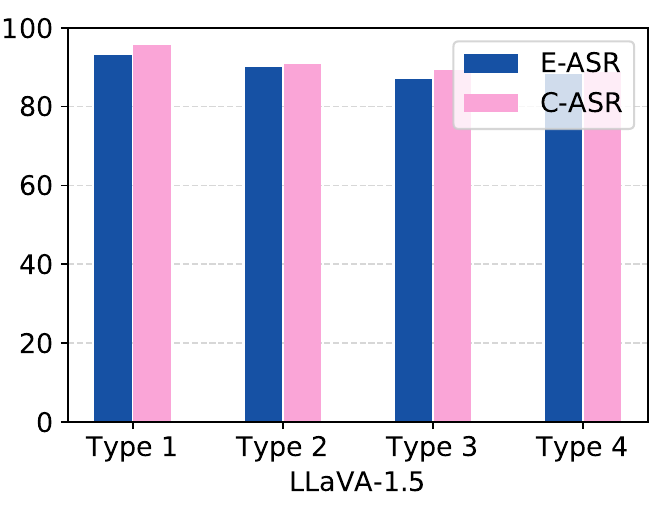}
    \hspace{-4pt}
    \includegraphics[width=0.246\textwidth]{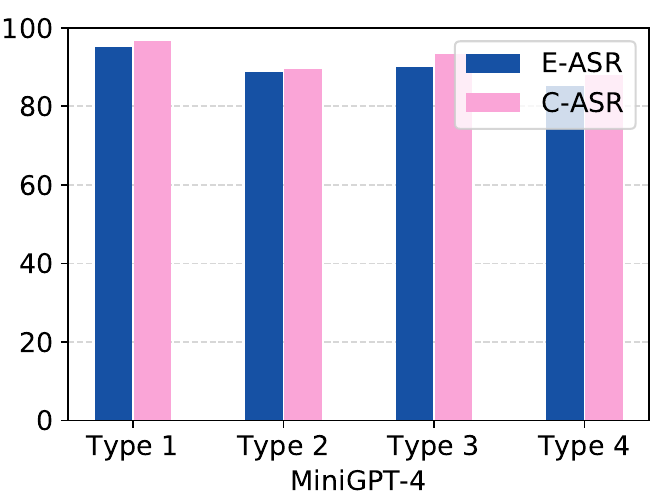}
    \hspace{-4pt}
    \includegraphics[width=0.246\textwidth]{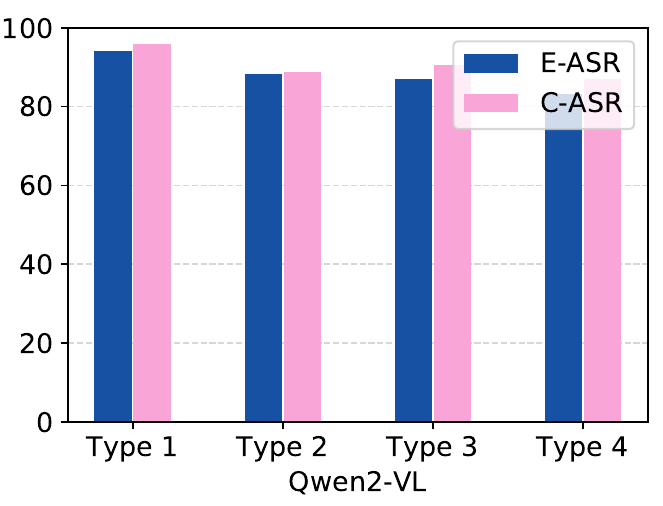}
    \hspace{-4pt}
    \includegraphics[width=0.246\textwidth]{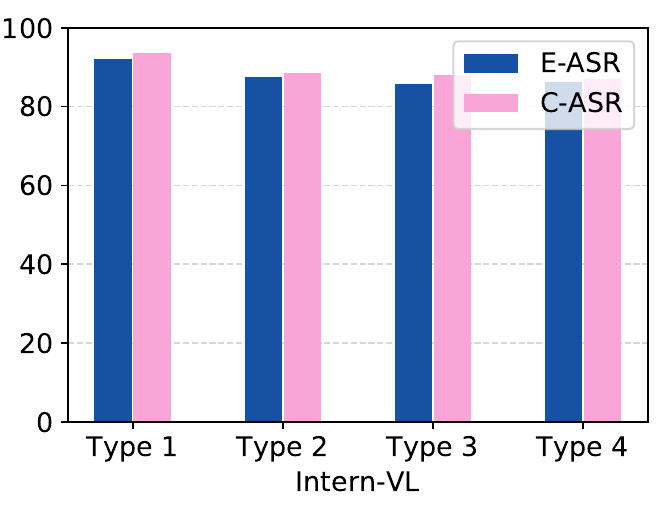}
    \vspace{-10pt}
    \caption{Attack performance with different attackers-chosen adversarial targets. Types 1-4 denote refusal  response, irrelevant semantic, misleading prefix, and toxic language, respectively.}
    \label{fig:ablation2}
\end{figure*}

\noindent \textbf{Evaluations of attacking on different targets.}
We explore the effectiveness of our method when simultaneously attacking different
targets. As shown in Figure~\ref{fig:ablation2}, we set four different adversarial targets: refusal response, irrelevant semantic, misleading prefix, and toxic language. Results show that our attack can achieve great attack performance across different adversarial targets, demonstrating the strong generalization ability and stability of our proposed attack. It also denotes that our attack is not sensitive to the attacker's goal and can be widely deployed in the real-world scenarios.

\noindent \textbf{Ablation on detailed designs.}
We provide detailed ablations on the in-depth design of our core components. As shown in Table~\ref{tab:ablation3}, we investigate the designs of both phase-aware perturbation and adversarial prompt enhancement. As for phase-aware designs, we replace our utilized DFT with other spectral tools DWT and DCT. We exploit their low- and high-frequency bands to denote the phase and amplitude spectrum. Results show that our utilized DFT achieves the best performance, because DFT can explicitly represent the phase-aware patterns representing the visual structure contexts. As for adversarial learning, leveraging auxiliary learnable prompts with original ones leads to better attack performance.

\begin{table}[t!]
    \centering
    \caption{Ablation of detailed designs on ImageNet dataset \cite{deng2009imagenet}.}
    \vspace{-6pt}
    \setlength{\tabcolsep}{1.0mm}{
    \begin{tabular}{c|c|ccc|ccc}
    \hline
    \multirow{2}*{Component} & \multirow{2}*{Variant} 
    & \multicolumn{3}{c|}{Qwen2-VL \cite{wang2024qwen2}} & \multicolumn{3}{c}{Intern-VL \cite{chen2024internvl}} \\
    ~ & ~ & SS & E-ASR & C-ASR & SS & E-ASR & C-ASR \\ \hline
    \multirow{3}*{\makecell[c]{Phase-aware\\Design}} & DWT & 0.876 & 86.4 & 87.0 & 0.858 & 84.9 & 85.7 \\
    ~ & DCT & 0.839 & 82.4 & 84.2 & 0.846 & 85.5 & 86.9 \\
    ~ & DFT & \textbf{0.895} & \textbf{88.3} & \textbf{88.7} 
                & \textbf{0.887} & \textbf{87.6} & \textbf{88.5} \\ \hline
    \multirow{3}*{\makecell[c]{Adversarial\\Prompt}} & origin & 0.815 & 81.8 & 83.2 & 0.811 & 80.6 & 82.7 \\
    ~ & auxiliary & 0.852 & 84.1 & 85.8 & 0.844 & 84.3 & 84.9 \\
    ~ & both & \textbf{0.895} & \textbf{88.3} & \textbf{88.7} 
                & \textbf{0.887} & \textbf{87.6} & \textbf{88.5} \\ \hline
    \end{tabular}}
    \label{tab:ablation3}
\end{table}

\begin{figure*}[t!]
    \centering
    \includegraphics[width=0.334\textwidth]{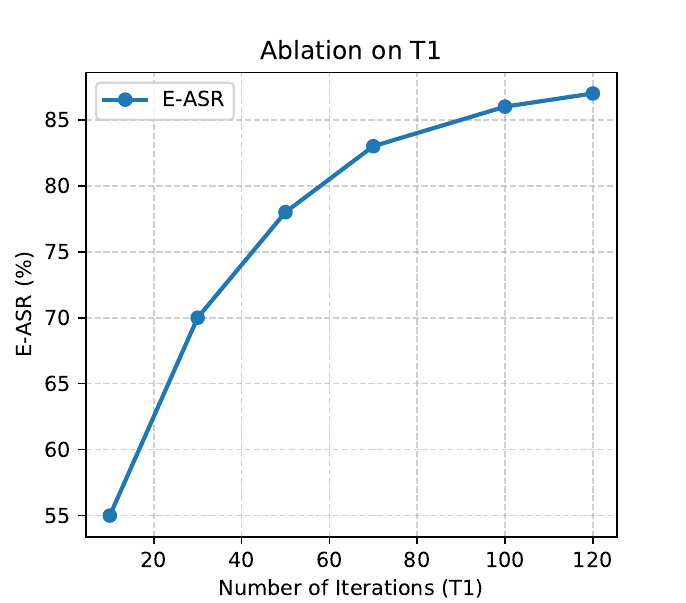}
    \hspace{-6pt}
    \includegraphics[width=0.326\textwidth]{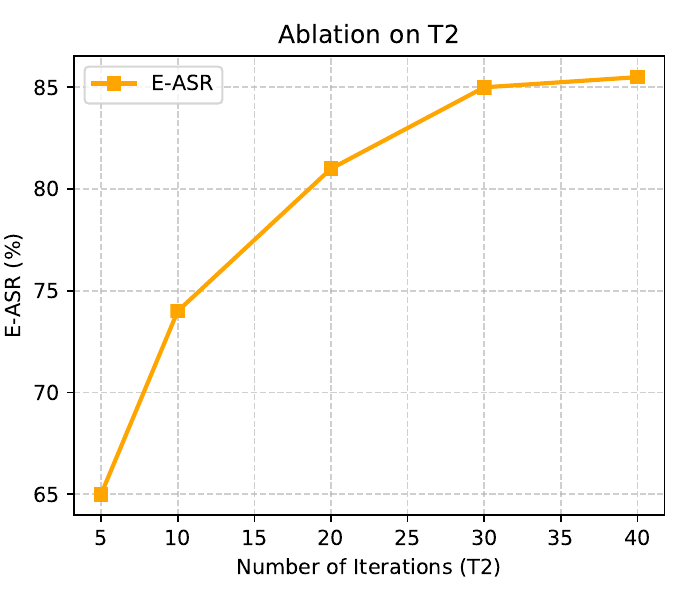}
    \hspace{-6pt}
    \includegraphics[width=0.326\textwidth]{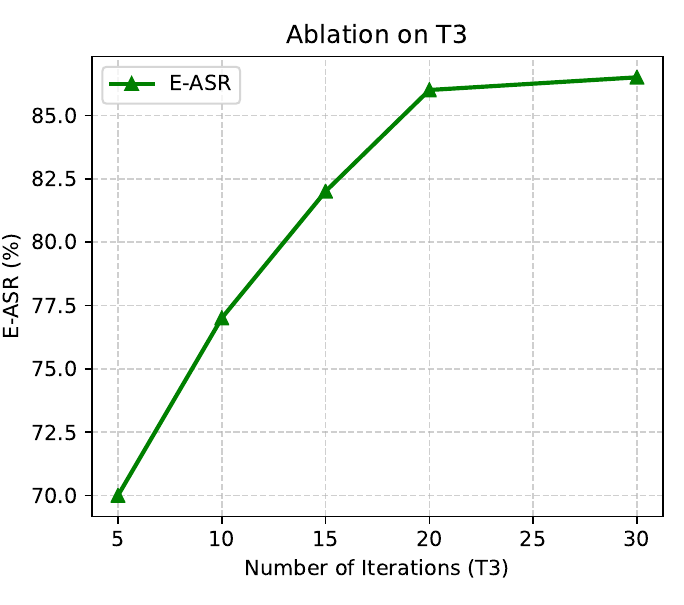}
    \vspace{-6pt}
    \caption{Ablation study on iteration numbers on BLIP-2.}
    \label{fig:ablation3}
\end{figure*}

\noindent \textbf{Ablation on iteration numbers.}
We provide the ablation of iteration numbers in Figure~\ref{fig:ablation3}. To take a balance between the attack performance and resource cost, we set $T_1=100,T_2=30,T_3=20$ in our experiments.

\noindent \textbf{\blue{{Cross-prompt stability analysis.}}}
\blue{To investigate whether the effectiveness of the generated perturbations depends on the auxiliary prompt, we evaluate adversarial examples under different inference prompts. Specifically, adversarial examples are generated using the combined prompt ($p+p_{aux}$), while evaluation is conducted using the original prompt, paraphrased prompts, and template-rewritten prompts.
As shown in Table~\ref{tab:prompt_stability}, removing the auxiliary prompt during inference only leads to a minor decrease in attack success rate. Similar behavior is observed under prompt paraphrasing and template modifications. These results suggest that the auxiliary prompt primarily serves as an optimization-time semantic guidance mechanism rather than a necessary attack component during deployment. The generated perturbations exhibit strong cross-prompt transferability and remain effective under the original prompt alone.
}

\begin{table}[t]
\small
\centering
\caption{\blue{Cross-prompt stability analysis of the auxiliary prompt.}}
\vspace{-6pt}
\begin{tabular}{l|ccc}
\hline
\blue{Inference Prompt}
&
\blue{SS}
&
\blue{E-ASR}
&
\blue{C-ASR}
\\
\hline

\blue{$p+p_{aux}$}
&
\blue{0.92}
&
\blue{89.7}
&
\blue{91.4}
\\

\blue{Original Prompt $p$}
&
\blue{0.91}
&
\blue{87.8}
&
\blue{89.9}
\\

\blue{Paraphrased Prompt}
&
\blue{0.90}
&
\blue{86.5}
&
\blue{88.4}
\\

\blue{Template-Rewritten Prompt}
&
\blue{0.90}
&
\blue{85.9}
&
\blue{87.6}
\\

\hline
\end{tabular}
\label{tab:prompt_stability}
\end{table}

\begin{table*}[t]
\centering
\caption{\blue{Robustness against stronger adaptive defenses on Qwen2-VL and ImageNet.}}
\vspace{-6pt}
\begin{tabular}{c|ccc|ccc|ccc|ccc}
\hline

\multirow{2}{*}{\blue{Attack}}
&
\multicolumn{3}{c|}{\blue{Adv. Fine-Tuning}}
&
\multicolumn{3}{c|}{\blue{Freq. Filtering}}
&
\multicolumn{3}{c|}{\blue{Rand. Transform}}
&
\multicolumn{3}{c}{\blue{Token Consistency Check}}
\\

&
\blue{SS} & \blue{E-ASR} & \blue{C-ASR}
&
\blue{SS} & \blue{E-ASR} & \blue{C-ASR}
&
\blue{SS} & \blue{E-ASR} & \blue{C-ASR}
&
\blue{SS} & \blue{E-ASR} & \blue{C-ASR}
\\

\hline

\blue{APGD \cite{cui2024robustness}}
&
\blue{0.412} & \blue{41.8} & \blue{43.2}
&
\blue{0.435} & \blue{43.7} & \blue{44.9}
&
\blue{0.446} & \blue{44.2} & \blue{45.8}
&
\blue{0.421} & \blue{42.1} & \blue{43.7}
\\

\blue{MF-Att \cite{zhao2024evaluating}}
&
\blue{0.468} & \blue{47.6} & \blue{49.2}
&
\blue{0.482} & \blue{48.8} & \blue{50.3}
&
\blue{0.491} & \blue{49.4} & \blue{51.2}
&
\blue{0.473} & \blue{47.9} & \blue{49.6}
\\

\blue{CroPA \cite{luo2024image}}
&
\blue{0.517} & \blue{51.9} & \blue{53.6}
&
\blue{0.532} & \blue{53.1} & \blue{54.7}
&
\blue{0.541} & \blue{54.3} & \blue{55.8}
&
\blue{0.522} & \blue{52.2} & \blue{53.9}
\\

\blue{MABA \cite{liang2024revisiting}}
&
\blue{0.593} & \blue{59.8} & \blue{61.5}
&
\blue{0.606} & \blue{60.7} & \blue{62.4}
&
\blue{0.615} & \blue{61.6} & \blue{63.8}
&
\blue{0.598} & \blue{60.1} & \blue{61.8}
\\

\blue{VMA \cite{wang2025attention}} 
&
\blue{0.574} & \blue{57.3} & \blue{59.8}
&
\blue{0.588} & \blue{58.6} & \blue{60.7}
&
\blue{0.601} & \blue{59.5} & \blue{61.8}
&
\blue{0.580} & \blue{57.9} & \blue{60.1}
\\

\blue{\textbf{Ours}}
&
\blue{\textbf{0.734}}
&
\blue{\textbf{73.6}}
&
\blue{\textbf{75.8}}
&
\blue{\textbf{0.761}}
&
\blue{\textbf{76.2}}
&
\blue{\textbf{78.1}}
&
\blue{\textbf{0.748}}
&
\blue{\textbf{74.8}}
&
\blue{\textbf{76.9}}
&
\blue{\textbf{0.722}}
&
\blue{\textbf{72.5}}
&
\blue{\textbf{74.6}}
\\

\hline
\end{tabular}
\label{tab:adaptive_defense}
\end{table*}

\subsection{More Analysis}

\begin{table}[t!]
    \centering
    \caption{Defense on Qwen2-VL \cite{wang2024qwen2} and ImageNet \cite{deng2009imagenet} dataset.}
    \vspace{-6pt}
    \setlength{\tabcolsep}{1.2mm}{
    \begin{tabular}{c|ccc|ccc}
    \hline
    \multirow{2}*{Attack} & \multicolumn{3}{c|}{With Fine-tuning} & \multicolumn{3}{c}{With Purification} \\
    ~ & SS & E-ASR & C-ASR & SS & E-ASR & C-ASR  \\ \hline
     APGD \cite{cui2024robustness} & 0.484 & 47.8 & 49.6 & 0.495 & 48.3 & 49.2\\
     MF-Att \cite{zhao2024evaluating} & 0.525 & 55.1 & 55.1 & 0.507 & 51.9 & 52.0 \\
     CroPA \cite{luo2024image} & 0.567 & 56.2 & 57.3 & 0.580 & 57.4 & 58.6 \\
     MABA \cite{liang2024revisiting} & 0.638 & 64.1 & 65.2 & 0.640 & 64.4 & 66.5 \\
     VMA \cite{wang2025attention} & 0.619 & 60.3 & 63.4 & 0.598 & 60.2 & 62.8 \\
     Ours & 0.786 & 78.1 & 80.6 & 0.835 & 83.7 & 83.7 \\ \hline
    \end{tabular}}
    \label{tab:defense1}
\end{table}

\begin{table}[t!]
    \centering
    \caption{Defense on Claude-3.5 \cite{anthropic2024claude} and ImageNet \cite{deng2009imagenet} dataset.}
    \vspace{-6pt}
    \setlength{\tabcolsep}{1.2mm}{
    \begin{tabular}{c|ccc|ccc}
    \hline
    \multirow{2}*{Attack} & \multicolumn{3}{c|}{With Fine-tuning} & \multicolumn{3}{c}{With Purification} \\
    ~ & SS & E-ASR & C-ASR & SS & E-ASR & C-ASR  \\ \hline
    MF-Att \cite{zhao2024evaluating} & 0.07 & 7.5 & 7.9 & 0.09 & 4.0 & 9.1 \\
    M-Att \cite{li2025frustratingly} & 0.16 & 14.8 & 14.8 & 0.20 & 15.9 & 17.6 \\
    FOA \cite{jia2025adversarial} & 0.25 & 22.7 & 24.3 & 0.23 & 22.5 & 23.9 \\
    Ours & 0.49 & 46.7 & 49.2 & 0.51 & 47.4 & 49.8\\
    \hline
    \end{tabular}}
    \label{tab:defense2}
\end{table}

\noindent \textbf{Robustness to potential defenses.}
To investigate the robustness of our proposed attack, we evaluate attack methods against potential defenses in a realistic white-box setting defense (\textit{i.e.}, fine-tuning \cite{zhu2023enhancing}, one of the most widely used methods) and a state-of-the-art black-box setting defense (\textit{i.e.}, zero-shot image purification \cite{shi2023black}). As shown in Table~\ref{tab:defense1} and Table~\ref{tab:defense2}, we report the performance on both open-source and closed-source MLLM models. Results show that our attack still achieves much better performance than the compared attack baselines, demonstrating that our method is more robust to defense strategies.
We also try to implement more defenses: Resizing, Smoothing Filters, and JPEG Compression as shown in Table \ref{tab:defense_r}, where our method is more robust.

\blue{To further investigate whether our attack remains effective against defenses specifically related to frequency-domain manipulations, we additionally evaluate four stronger adaptive
defenses, including adversarial fine-tuning, frequency-domain
filtering, randomized transformations, and visual-token
consistency checking.
As shown in Table~\ref{tab:adaptive_defense}, these defenses reduce attack success
rates for all methods. Nevertheless, our attack consistently
achieves the highest E-ASR and C-ASR among all baselines.
This suggests that the effectiveness of our method does not
solely originate from removable high-frequency artifacts,
but is closely related to phase-sensitive semantic structures.}

\begin{table}[t!]
    \centering
    \caption{Defense on Qwen2-VL and ImageNet dataset.}
    \vspace{-6pt}
    \setlength{\tabcolsep}{0.8mm}{
    \begin{tabular}{c|cccccc}
    \hline
    		Defense &	      APGD  & MF-Att  & CroPA  & MABA  & VMA  & Ours \\ \hline
Resizing        &    0.462  &0.481  &0.497  &0.536 & 0.528  &0.824\\ 
Smoothing Filters   & 0.489  &0.505  &0.522  &0.563  &0.570 & 0.797\\ 
JPEG Compression  & 0.523  &0.538 & 0.556 & 0.604  &0.595 & 0.781\\ \hline

    \end{tabular}}
    \label{tab:defense_r}
\end{table}

\begin{figure}[t!]
    \centering
    \includegraphics[width=0.48\textwidth]{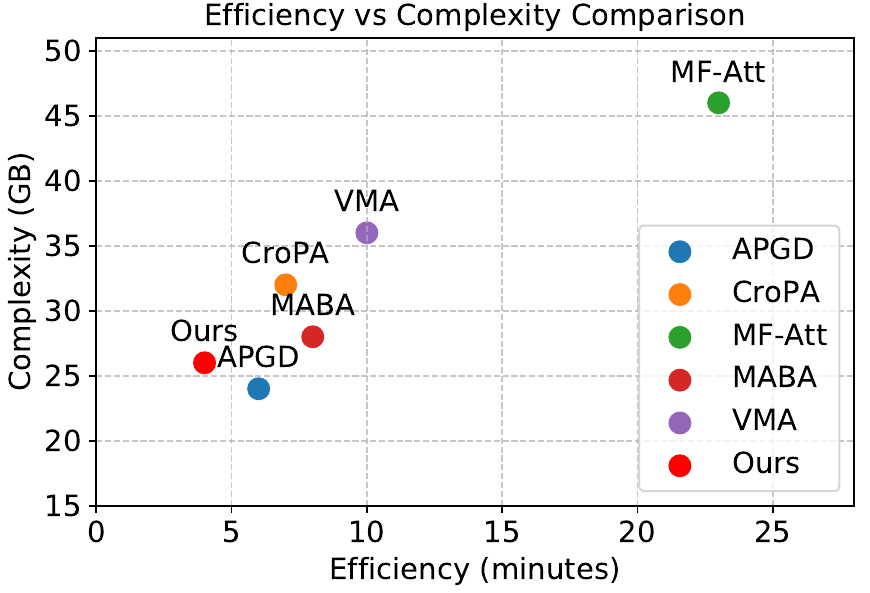}
    \caption{Comparisons on attacks' efficiency/complexity.}
    \label{fig:efficiency}
\end{figure}

\noindent \textbf{Complexity and efficiency.}
To investigate the attacks' practicability, we also provide the analysis of attacks' efficiency and complexity on single adversarial example generation. As shown in Figure~\ref{fig:efficiency}, our attack has very competitive efficiency and complexity compared with other MLLM attacks while achieving much better attack performance, demonstrating our practicability.

\begin{figure*}[t!]
    \centering
    \includegraphics[width=0.96\textwidth]{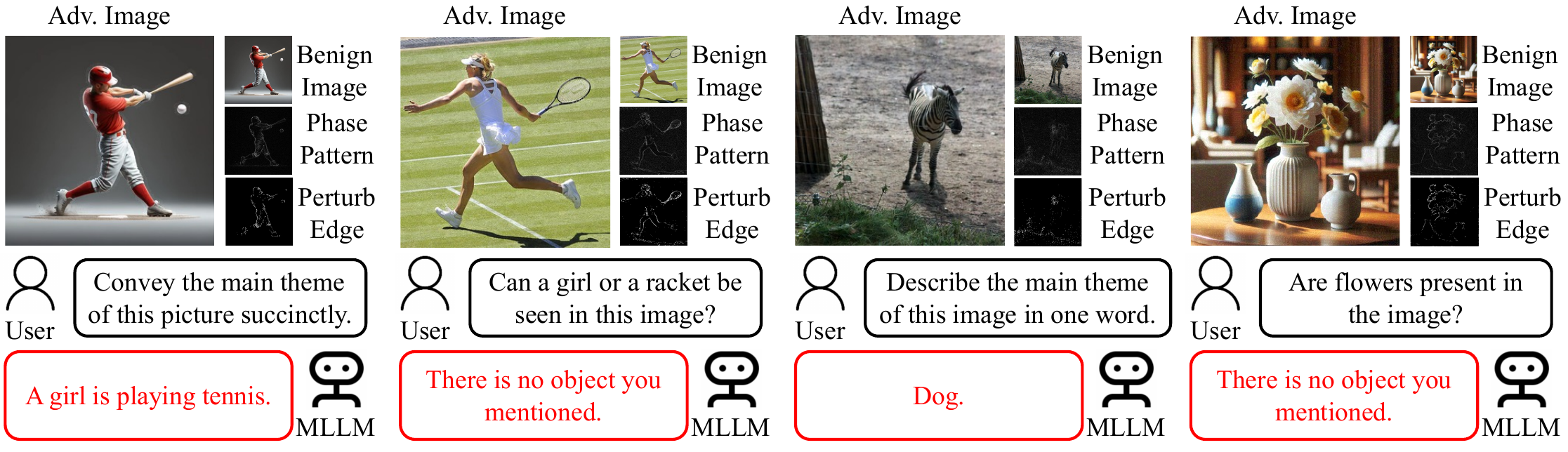}
    \caption{Visualization results of our proposed phase-aware MLLM attack.}
    \label{fig:result}
\end{figure*}

\begin{table}[t!]
\centering
\caption{Transfer attack performance on ImageNet.}
\vspace{-6pt}
\label{tab:transfer_attack}
\setlength{\tabcolsep}{1.6mm}{
\begin{tabular}{lccc|ccc}
\hline
\multirow{2}*{Attack}& \multicolumn{3}{c}{BLIP-2 $\rightarrow$ Qwen2-VL}
& \multicolumn{3}{c}{BLIP-2 $\rightarrow$ Intern-VL} \\
\cline{2-4} \cline{5-7}
 & SS & E-ASR & C-ASR & SS & E-ASR & C-ASR \\
\hline
APGD \cite{cui2024robustness}   & 0.582 & 50.2 & 52.6 & 0.565 & 47.8 & 50.1 \\
MF-Att \cite{zhao2024evaluating} & 0.626 & 55.4 & 57.9 & 0.607 & 52.8 & 55.3 \\
CroPA \cite{luo2024image}  & 0.651 & 58.9 & 61.2 & 0.634 & 56.5 & 58.7 \\
MABA \cite{liang2024revisiting}   & 0.693 & 64.3 & 66.8 & 0.679 & 61.7 & 63.9 \\
VMA \cite{wang2025attention}    & 0.718 & 66.9 & 69.1 & 0.692 & 64.2 & 66.5 \\
\hline
Ours & \textbf{0.783} & \textbf{74.8} & \textbf{77.2} 
              & \textbf{0.765} & \textbf{71.3} & \textbf{73.9} \\
\hline
\end{tabular}}
\end{table}

\noindent \textbf{Transfer attack.} We try to implement transfer-attack comparison in Table~\ref{tab:transfer_attack}, where we generate adversarial examples on BLIP-2 and test them on other MLLM models. The results show that our attack still achieves better performance, demonstrating our effectiveness.

\begin{table}[t]
\centering
\caption{LLM-based evaluation.}
\vspace{-6pt}
\label{tab:gpt_assessment}
\setlength{\tabcolsep}{0.6mm}{
\begin{tabular}{ccccccc}
\hline
\textbf{Attack} & \textbf{BLIP-2} & \textbf{LLaVA-1.5} & \textbf{Flamingo} & \textbf{MiniGPT-4} & \textbf{Qwen2-VL} & \textbf{Intern-VL} \\
\hline
APGD   & 0.41 & 0.38 & 0.35 & 0.37 & 0.40 & 0.39 \\
MF-Att & 0.47 & 0.44 & 0.41 & 0.43 & 0.46 & 0.45 \\
CroPA  & 0.51 & 0.48 & 0.45 & 0.47 & 0.50 & 0.49 \\
MABA   & 0.57 & 0.54 & 0.51 & 0.53 & 0.56 & 0.55 \\
VMA    & 0.60 & 0.57 & 0.54 & 0.56 & 0.59 & 0.58 \\
\hline
\textbf{Ours} & \textbf{0.68} & \textbf{0.65} & \textbf{0.62} & \textbf{0.64} & \textbf{0.67} & \textbf{0.66} \\
\hline
\end{tabular}}
\end{table}

\noindent \textbf{LLM-as-Judge.} To investigate whether our attack is sensitive to the judge model, we replace the text evaluator with the LLM model. We utilize LLM-based assessment (GPT-4o) to evaluate the similarity in the range of [0,1] as shown in Table \ref{tab:gpt_assessment}. \blue{Given a generated response and a target response, GPT-4o evaluates whether the generated output semantically expresses the target content regardless of wording variations. The evaluation uses deterministic decoding (temperature = 0) and a fixed prompt template for all methods.} It demonstrates that utilizing LLM to evaluate the textual output is also a good solution.

\begin{table}[t!]
\centering
\caption{\blue{Imperceptibility comparison of different MLLM attack methods.}}
\vspace{-6pt}
\resizebox{0.5\textwidth}{!}{
\setlength{\tabcolsep}{0.6mm}{
\begin{tabular}{c|ccccccc}
\hline
\blue{Method}
&
\blue{LPIPS$\downarrow$}
&
\blue{SSIM$\uparrow$}
&
\blue{PSNR$\uparrow$}
&
\blue{$L_0(\%)\downarrow$}
&
\blue{$L_2\downarrow$}
&
\blue{$L_\infty\downarrow$}
&
\blue{Mask Ratio(\%)$\downarrow$}
\\
\hline

\blue{MF-Att}
&
\blue{0.086}
&
\blue{0.974}
&
\blue{38.6}
&
\blue{100.0}
&
\blue{7.34}
&
\blue{8}
&
\blue{100.0}
\\

\blue{M-Att}
&
\blue{0.074}
&
\blue{0.978}
&
\blue{39.8}
&
\blue{100.0}
&
\blue{6.81}
&
\blue{8}
&
\blue{100.0}
\\

\blue{M-Att v2}
&
\blue{0.068}
&
\blue{0.981}
&
\blue{40.9}
&
\blue{100.0}
&
\blue{6.17}
&
\blue{8}
&
\blue{100.0}
\\

\blue{FOA}
&
\blue{0.061}
&
\blue{0.984}
&
\blue{41.7}
&
\blue{100.0}
&
\blue{5.64}
&
\blue{8}
&
\blue{100.0}
\\

\blue{Ours}
&
\blue{\textbf{0.045}}
&
\blue{\textbf{0.991}}
&
\blue{\textbf{44.3}}
&
\blue{\textbf{24.7}}
&
\blue{\textbf{3.92}}
&
\blue{8}
&
\blue{\textbf{21.3}}
\\

\hline
\end{tabular}}}
\label{tab:imperceptibility}
\end{table}

\noindent \textbf{\blue{Perceptual quality analysis.}}
\blue{As shown in Table~\ref{tab:imperceptibility}, although all methods are evaluated under the same perturbation budget ($L_\infty=8/255$), the proposed phase-aware localization strategy significantly reduces the perturbed area ratio from 100\% to only 21.3\%, leading to substantially better perceptual quality in terms of LPIPS, SSIM, and PSNR.}

\subsection{Visualization}
We provide visualization results of our generated adversarial examples as shown in Figure~\ref{fig:result}. Here, we provide the benign image, the decomposed phase pattern, the phase-aware edge region, and the final adversarial images. Results show that our attack can effectively fool MLLMs by suppressing their intrinsic phase focus.

\subsection{Discussion on Limitations and Failure Cases}

\new{Although the proposed phase-aware attack framework achieves strong performance across diverse multimodal large language models (MLLMs), its effectiveness relies on the assumption that phase components consistently encode semantically meaningful structural cues. This assumption may not fully hold under certain visual conditions, leading to degraded attack performance.
\textit{(1) Complex and cluttered scenes.}
In highly cluttered scenes containing multiple objects or dense backgrounds, the phase spectrum tends to encode a mixture of overlapping structural patterns. As a result, the extracted phase-dominant regions become less spatially localized and may include redundant or irrelevant structures. This reduces the accuracy of phase-aware masking and leads to suboptimal perturbation allocation, thereby weakening attack effectiveness.
\textit{(2) Weak structural or low-contrast images.}
Our method relies on the stability of edge- and contour-level information in the phase domain. However, for images with weak edges, motion blur, or low object-background contrast, the phase spectrum does not provide sufficiently discriminative structural cues. In such cases, the derived phase-aware mask may fail to accurately reflect semantic regions, limiting the effectiveness of structure-guided perturbations.
\textit{(3) Small object dominance.}
When target objects occupy only a very small portion of the image, the available phase-sensitive region is inherently limited. Since our optimization constrains perturbations within phase-aware masks, the effective attack budget on semantically meaningful regions is reduced, leading to weaker adversarial impact and lower success rates.
\textit{(4) Text-rich and OCR-centric scenarios.}
For images containing dense textual content or requiring optical character recognition (OCR), semantic understanding depends more on high-frequency character-level details rather than global structural information. Since our method primarily exploits phase-domain structural priors, it may not fully capture fine-grained textual semantics, resulting in reduced attack effectiveness in such scenarios.
\textit{(5) Fine-grained reasoning tasks.}
Some multimodal reasoning tasks require attribute-level discrimination, numerical reasoning, or spatial relationship inference beyond global structural cues. In these cases, phase representations alone may be insufficient to fully characterize the reasoning-critical visual features, leading to less pronounced improvements compared to standard recognition or captioning tasks.
Overall, these limitations suggest that while phase-aware perturbations effectively exploit structural sensitivity in MLLMs, the approach is inherently constrained by the reliability of phase-domain semantic alignment. Future work may benefit from integrating object-level segmentation, text-aware modeling, and hybrid frequency–spatial representations to improve robustness across more challenging visual distributions.}

\subsection{\new{Broader Applicability of Phase-aware Structural Modeling}}
\new{
Although this work focuses on adversarial attacks against multimodal large language models (MLLMs), the underlying principle of exploiting phase-aware structural representations is not limited to vision-language reasoning. Since phase information naturally captures object boundaries, spatial layouts, and semantic structures, the proposed frequency-aware modeling strategy has the potential to benefit a broader range of vision and multimodal learning problems.
\textit{First}, phase-aware structural representations could improve visual understanding tasks that require robust structural reasoning. For example, group activity recognition and compositional action recognition rely heavily on modeling interactions among multiple objects and their spatial configurations rather than solely appearance information. Recent studies have demonstrated the effectiveness of hierarchical graph reasoning and progressive feature learning for capturing such structured semantics~\cite{yan2020higcin,yan2023progressive,qu2026spatio}. Integrating phase-aware structural priors into these frameworks may provide complementary geometric cues for more robust representation learning.
\textit{Second}, our phase-guided localization mechanism may naturally extend to efficient visual token selection in large vision-language models. Existing token compression methods aim to preserve semantically important visual tokens while reducing computational overhead~\cite{xing2026vision}. Since the proposed phase-aware masks explicitly identify structure-sensitive regions that dominate semantic perception, they could provide an interpretable criterion for adaptive token pruning and visual token allocation.
\textit{Third}, the proposed structural perturbation paradigm is closely related to fine-grained cross-modal correspondence learning. Applications such as dense audio-visual event localization require accurate alignment between visual structures and semantic events~\cite{xing2025locality}. Phase-aware representations could provide additional structural consistency constraints to improve multimodal correspondence under challenging scenarios.
\textit{Finally}, our framework may also benefit few-shot visual recognition tasks, where learning transferable structural representations is often more important than modeling appearance variations. Recent few-shot recognition methods progressively align semantic features across different visual distributions~\cite{qu2025mvp,qu2026spatio}. Since phase information is relatively invariant to texture and illumination changes, incorporating phase-aware structural modeling may further improve feature transferability under limited supervision.
Overall, our work suggests that frequency-domain structural modeling is a general visual representation principle rather than a task-specific adversarial strategy. We believe that combining phase-aware semantic representations with graph reasoning, efficient token compression, cross-modal correspondence learning, and few-shot representation learning constitutes a promising direction for future research.}


\section{Conclusion}
In this paper, we investigate the intrinsic sensitivity of multimodal large language models (MLLMs) to different frequency components of visual inputs. Motivated by the human visual system and empirical observations, we find that MLLMs rely heavily on the phase spectrum, which encodes structural and semantic information. Based on this insight, we propose a novel phase-aware adversarial attack framework that localizes perturbations to phase-dominant regions and explicitly manipulates spectral semantics. To further amplify the adversarial impact, we design an auxiliary prompt module with the adversarial learning scheme to guide the model's attention during optimization. Extensive experiments across multiple MLLMs and tasks demonstrate the effectiveness and interpretability of our method.
This work highlights the importance of understanding MLLMs' frequency-domain behavior and opens up new directions for designing more principled and structure-aware adversarial strategies. Future work may explore how such frequency-level insights can benefit defense, detection, and robust training in multimodal systems.

\bibliographystyle{IEEEtran}
\bibliography{sample-base}

\begin{thebibliography}{10}
\providecommand{\url}[1]{#1}
\csname url@samestyle\endcsname
\providecommand{\newblock}{\relax}
\providecommand{\bibinfo}[2]{#2}
\providecommand{\BIBentrySTDinterwordspacing}{\spaceskip=0pt\relax}
\providecommand{\BIBentryALTinterwordstretchfactor}{4}
\providecommand{\BIBentryALTinterwordspacing}{\spaceskip=\fontdimen2\font plus
\BIBentryALTinterwordstretchfactor\fontdimen3\font minus \fontdimen4\font\relax}
\providecommand{\BIBforeignlanguage}[2]{{%
\expandafter\ifx\csname l@#1\endcsname\relax
\typeout{** WARNING: IEEEtran.bst: No hyphenation pattern has been}%
\typeout{** loaded for the language `#1'. Using the pattern for}%
\typeout{** the default language instead.}%
\else
\language=\csname l@#1\endcsname
\fi
#2}}
\providecommand{\BIBdecl}{\relax}
\BIBdecl

\bibitem{li2023blip}
J.~Li, D.~Li, S.~Savarese, and S.~Hoi, ``Blip-2: Bootstrapping language-image pre-training with frozen image encoders and large language models,'' in \emph{International conference on machine learning}.\hskip 1em plus 0.5em minus 0.4em\relax PMLR, 2023, pp. 19\,730--19\,742.

\bibitem{liu2024visual}
H.~Liu, C.~Li, Q.~Wu, and Y.~J. Lee, ``Visual instruction tuning,'' \emph{Advances in neural information processing systems}, vol.~36, 2024.

\bibitem{xu2026ltkt}
J.~Xu, R.~Tang, P.~Lv, M.~Yu, G.~Yu, and E.~Chen, ``Ltkt: Knowledge tracing based on positive and negative learning transfers,'' \emph{Tsinghua Science and Technology}, vol.~31, no.~3, pp. 1894--1917, 2026.

\bibitem{alayrac2022flamingo}
J.-B. Alayrac, J.~Donahue, P.~Luc, A.~Miech, I.~Barr, Y.~Hasson, K.~Lenc, A.~Mensch, K.~Millican, M.~Reynolds \emph{et~al.}, ``Flamingo: a visual language model for few-shot learning,'' \emph{Advances in neural information processing systems}, vol.~35, pp. 23\,716--23\,736, 2022.

\bibitem{ye2026predicting}
L.~Ye, Z.-h. Li, H.-n. Yan, C.~Liu, H.~H. Cho, and T.~Guo, ``Predicting film-cooling effectiveness of compound-angle holes using a pod-based hybrid deep learning model,'' \emph{Aerospace Science and Technology}, p. 112590, 2026.

\bibitem{liu2021context}
D.~Liu, X.~Qu, J.~Dong, P.~Zhou, Y.~Cheng, W.~Wei, Z.~Xu, and Y.~Xie, ``Context-aware biaffine localizing network for temporal sentence grounding,'' in \emph{Proceedings of the IEEE/CVF Conference on Computer Vision and Pattern Recognition}, 2021, pp. 11\,235--11\,244.

\bibitem{liu2020jointly}
D.~Liu, X.~Qu, X.-Y. Liu, J.~Dong, P.~Zhou, and Z.~Xu, ``Jointly cross-and self-modal graph attention network for query-based moment localization,'' in \emph{Proceedings of the 28th ACM International Conference on Multimedia}, 2020, pp. 4070--4078.

\bibitem{liu2022memory}
D.~Liu, X.~Qu, X.~Di, Y.~Cheng, Z.~Xu, and P.~Zhou, ``Memory-guided semantic learning network for temporal sentence grounding,'' in \emph{Proceedings of the AAAI Conference on Artificial Intelligence}, vol.~36, no.~2, 2022, pp. 1665--1673.

\bibitem{jiang2025novel}
R.~Jiang, S.~Tong, J.~Wu, H.~Hu, R.~Zhang, H.~Wang, Y.~Zhao, W.~Zhu, S.~Li, and X.~Zhang, ``A novel eeg artifact removal algorithm based on an advanced attention mechanism,'' \emph{Scientific Reports}, vol.~15, no.~1, p. 19419, 2025.

\bibitem{liu2021adaptive}
D.~Liu, X.~Qu, J.~Dong, and P.~Zhou, ``Adaptive proposal generation network for temporal sentence localization in videos,'' in \emph{Proceedings of the 2021 Conference on Empirical Methods in Natural Language Processing}, 2021, pp. 9292--9301.

\bibitem{liu2022unsupervised}
D.~Liu, X.~Qu, Y.~Wang, X.~Di, K.~Zou, Y.~Cheng, Z.~Xu, and P.~Zhou, ``Unsupervised temporal video grounding with deep semantic clustering,'' in \emph{Proceedings of the AAAI Conference on Artificial Intelligence}, vol.~36, no.~2, 2022, pp. 1683--1691.

\bibitem{liu2022reducing}
D.~Liu, X.~Qu, and W.~Hu, ``Reducing the vision and language bias for temporal sentence grounding,'' in \emph{Proceedings of the 30th ACM International Conference on Multimedia}, 2022, pp. 4092--4101.

\bibitem{wang2025giving}
L.~Wang, Y.~Ma, Z.~Yan, L.~Zhang, Y.~Hu, and S.~Zhao, ``Giving or receiving: Impact of online socializing in online fitness community on physical activity and emotional state,'' \emph{Computers in Human Behavior}, vol. 169, p. 108669, 2025.

\bibitem{lv2026adaptive}
J.~Lv, C.~Wang, and L.~Xie, ``Adaptive distributed observer design for nonlinear multiagent systems,'' \emph{Automatica}, vol. 183, p. 112625, 2026.

\bibitem{rombach2022high}
R.~Rombach, A.~Blattmann, D.~Lorenz, P.~Esser, and B.~Ommer, ``High-resolution image synthesis with latent diffusion models,'' in \emph{Proceedings of the IEEE/CVF conference on computer vision and pattern recognition}, 2022, pp. 10\,684--10\,695.

\bibitem{chen2024vision}
Y.~Chen, T.~He, J.~Fu, L.~Wang, J.~Guo, T.~Hu, and H.~Cheng, ``Vision-language meets the skeleton: Progressively distillation with cross-modal knowledge for 3d action representation learning,'' \emph{IEEE Transactions on Multimedia}, 2024.

\bibitem{qi2025vision}
Y.~Qi, H.~Li, Y.~Song, X.~Wu, and J.~Luo, ``How vision-language tasks benefit from large pre-trained models: A survey,'' \emph{IEEE Transactions on Multimedia}, 2025.

\bibitem{brown2020language}
T.~Brown, B.~Mann, N.~Ryder, M.~Subbiah, J.~D. Kaplan, P.~Dhariwal, A.~Neelakantan, P.~Shyam, G.~Sastry, A.~Askell \emph{et~al.}, ``Language models are few-shot learners,'' \emph{Advances in neural information processing systems}, vol.~33, pp. 1877--1901, 2020.

\bibitem{touvron2023llama}
H.~Touvron, T.~Lavril, G.~Izacard, X.~Martinet, M.-A. Lachaux, T.~Lacroix, B.~Rozi{\`e}re, N.~Goyal, E.~Hambro, F.~Azhar \emph{et~al.}, ``Llama: Open and efficient foundation language models,'' \emph{arXiv preprint arXiv:2302.13971}, 2023.

\bibitem{touvron2023llama2}
H.~Touvron, L.~Martin, K.~Stone, P.~Albert, A.~Almahairi, Y.~Babaei, N.~Bashlykov, S.~Batra, P.~Bhargava, S.~Bhosale \emph{et~al.}, ``Llama 2: Open foundation and fine-tuned chat models,'' \emph{arXiv preprint arXiv:2307.09288}, 2023.

\bibitem{liu2024survey}
D.~Liu, M.~Yang, X.~Qu, P.~Zhou, W.~Hu, and Y.~Cheng, ``A survey of attacks on large vision-language models: Resources, advances, and future trends,'' \emph{arXiv preprint arXiv:2407.07403}, 2024.

\bibitem{wang2025invisible}
B.~Wang, S.~Qian, and C.~Xu, ``Invisible backdoor attack with siamese tuning on pre-trained vision-language models,'' \emph{IEEE Transactions on Multimedia}, 2025.

\bibitem{hu2026trajectory}
M.~Hu, Z.~Li, H.~Wang, Y.~Wang, Y.~Ding, C.~Liu, and Q.~Song, ``Trajectory-aware attack: Explainable adversarial attack against multiple object trackers,'' \emph{IEEE Transactions on Multimedia}, 2026.

\bibitem{wang2025exploring}
Y.~Wang, W.~Hu, Y.~Dong, H.~Zhang, H.~Su, and R.~Hong, ``Exploring transferability of multimodal adversarial samples for vision-language pre-training models with contrastive learning,'' \emph{IEEE Transactions on Multimedia}, 2025.

\bibitem{liu2024pandora}
D.~Liu, M.~Yang, X.~Qu, P.~Zhou, X.~Fang, K.~Tang, Y.~Wan, and L.~Sun, ``Pandora's box: Towards building universal attackers against real-world large vision-language models,'' in \emph{The Thirty-eighth Annual Conference on Neural Information Processing Systems}, 2024.

\bibitem{yan2025fit}
H.~Yan, H.~Ma, X.~Cai, D.~Liu, Z.~Yuan, X.~Qu, J.~Dong, R.~Guan, X.~Fang, H.~He \emph{et~al.}, ``Fit the distribution: Cross-image/prompt adversarial attacks on multimodal large language models,'' in \emph{The Thirty-ninth Annual Conference on Neural Information Processing Systems}, 2025.

\bibitem{cai2025towards}
X.~Cai, D.~Liu, X.~Qu, X.~Fang, J.~Dong, K.~Tang, P.~Zhou, L.~Sun, and W.~Hu, ``Towards building model/prompt-transferable attackers against large vision-language models,'' in \emph{The Thirty-ninth Annual Conference on Neural Information Processing Systems}, 2025.

\bibitem{liu2026spatial}
D.~Liu, B.~Chen, and W.~Hu, ``Spatial-spectral homogeneous attacks on physical-world large vision-language models,'' in \emph{Proceedings of the AAAI Conference on Artificial Intelligence}, vol.~40, no.~9, 2026, pp. 7114--7122.

\bibitem{liu2026large}
D.~Liu, X.~Cai, P.~Zhou, X.~Qu, L.~Sun, and W.~Hu, ``Are large vision-language models robust to adversarial visual transformations?'' \emph{IEEE Transactions on Information Forensics and Security}, 2026.

\bibitem{liu2026generating}
D.~Liu, W.~Liu, X.~Cai, P.~Zhou, R.~Guan, X.~Qu, and B.~Du, ``Generating transferable attacks across large vision-language models using adversarial deformation learning,'' \emph{Pattern Recognition}, p. 113194, 2026.

\bibitem{shayegani2023jailbreak}
E.~Shayegani, Y.~Dong, and N.~Abu-Ghazaleh, ``Jailbreak in pieces: Compositional adversarial attacks on multi-modal language models,'' in \emph{The Twelfth International Conference on Learning Representations}, 2023.

\bibitem{bailey2023image}
L.~Bailey, E.~Ong, S.~Russell, and S.~Emmons, ``Image hijacks: Adversarial images can control generative models at runtime,'' \emph{arXiv preprint arXiv:2309.00236}, 2023.

\bibitem{dong2023robust}
Y.~Dong, H.~Chen, J.~Chen, Z.~Fang, X.~Yang, Y.~Zhang, Y.~Tian, H.~Su, and J.~Zhu, ``How robust is google's bard to adversarial image attacks?'' \emph{arXiv preprint arXiv:2309.11751}, 2023.

\bibitem{wang2023instructta}
X.~Wang, Z.~Ji, P.~Ma, Z.~Li, and S.~Wang, ``Instructta: Instruction-tuned targeted attack for large vision-language models,'' \emph{arXiv preprint arXiv:2312.01886}, 2023.

\bibitem{wang2024stop}
Z.~Wang, Z.~Han, S.~Chen, F.~Xue, Z.~Ding, X.~Xiao, V.~Tresp, P.~Torr, and J.~Gu, ``Stop reasoning! when multimodal llms with chain-of-thought reasoning meets adversarial images,'' \emph{arXiv preprint arXiv:2402.14899}, 2024.

\bibitem{zhang2024avibench}
H.~Zhang, W.~Shao, H.~Liu, Y.~Ma, P.~Luo, Y.~Qiao, and K.~Zhang, ``Avibench: Towards evaluating the robustness of large vision-language model on adversarial visual-instructions,'' \emph{arXiv preprint arXiv:2403.09346}, 2024.

\bibitem{lu2024test}
D.~Lu, T.~Pang, C.~Du, Q.~Liu, X.~Yang, and M.~Lin, ``Test-time backdoor attacks on multimodal large language models,'' \emph{arXiv preprint arXiv:2402.08577}, 2024.

\bibitem{luo2024image}
H.~Luo, J.~Gu, F.~Liu, and P.~Torr, ``An image is worth 1000 lies: Adversarial transferability across prompts on vision-language models,'' \emph{arXiv preprint arXiv:2403.09766}, 2024.

\bibitem{tao2024imgtrojan}
X.~Tao, S.~Zhong, L.~Li, Q.~Liu, and L.~Kong, ``Imgtrojan: Jailbreaking vision-language models with one image,'' \emph{arXiv preprint arXiv:2403.02910}, 2024.

\bibitem{zhao2024evaluating}
Y.~Zhao, T.~Pang, C.~Du, X.~Yang, C.~Li, N.-M.~M. Cheung, and M.~Lin, ``On evaluating adversarial robustness of large vision-language models,'' \emph{Advances in Neural Information Processing Systems}, vol.~36, 2024.

\bibitem{oppenheim2005importance}
A.~V. Oppenheim and J.~S. Lim, ``The importance of phase in signals,'' \emph{Proceedings of the IEEE}, vol.~69, no.~5, pp. 529--541, 2005.

\bibitem{karout2007two}
S.~Karout, \emph{Two-dimensional phase unwrapping}.\hskip 1em plus 0.5em minus 0.4em\relax Liverpool John Moores University (United Kingdom), 2007.

\bibitem{guo2008spatio}
C.~Guo, Q.~Ma, and L.~Zhang, ``Spatio-temporal saliency detection using phase spectrum of quaternion fourier transform,'' in \emph{2008 IEEE conference on computer vision and pattern recognition}.\hskip 1em plus 0.5em minus 0.4em\relax IEEE, 2008, pp. 1--8.

\bibitem{li2015finding}
J.~Li, L.-Y. Duan, X.~Chen, T.~Huang, and Y.~Tian, ``Finding the secret of image saliency in the frequency domain,'' \emph{IEEE transactions on pattern analysis and machine intelligence}, vol.~37, no.~12, pp. 2428--2440, 2015.

\bibitem{chen2021amplitude}
G.~Chen, P.~Peng, L.~Ma, J.~Li, L.~Du, and Y.~Tian, ``Amplitude-phase recombination: Rethinking robustness of convolutional neural networks in frequency domain,'' in \emph{Proceedings of the IEEE/CVF international conference on computer vision}, 2021, pp. 458--467.

\bibitem{qian2026individual}
Y.~Qian, Y.~Kong, Q.~Bao, Z.~Gu, B.~Wang, S.~Ji, J.~Zhang, and Z.~Lei, ``Individual \& common attack: Enhancing transferability in vlp models through modal feature exploitation,'' \emph{IEEE Transactions on Image Processing}, 2026.

\bibitem{qian2025exploiting}
Y.~Qian, X.~Zhu, Q.~Bao, F.~Yu, S.~Ji, Z.~Gu, W.~Wang, B.~Wang, and Z.~Lei, ``Exploiting shared adversarial features for dynamic attacks in large vision-language models,'' \emph{IEEE Transactions on Information Forensics and Security}, vol.~21, pp. 592--607, 2025.

\bibitem{qian2025multimodal}
Y.~Qian, Q.~Yu, Q.~Bao, S.~Ji, W.~Wang, B.~Wang, Z.~Gu, and Z.~Lei, ``A multimodal adversarial attack method via frequency domain enhancement and fine-grained cross-modal guidance,'' \emph{IEEE Transactions on Dependable and Secure Computing}, 2025.

\bibitem{lai2025enhancing}
C.-H. Lai, T.-E. Wu, and C.-C. Wang, ``Enhancing information security in smart manufacturing through least significant bit steganography in engineering drawings,'' \emph{Journal of Computing and Information Science in Engineering}, vol.~25, no.~9, p. 091006, 2025.

\bibitem{fan2025underwater}
R.~Fan, A.~Boukerche, P.~Pan, Z.~Jin, and Y.~Su, ``An underwater secure localization scheme based on physical layer cryptographic learning,'' \emph{IEEE Transactions on Mobile Computing}, 2025.

\bibitem{zhu2026data}
W.~Zhu, B.~Xu, B.~Guo, C.~Xie, and R.~Xiong, ``Data-efficient and interpretable long-term prognostics of pemfc degradation via physics-constrained symbolic regression--physics-informed neural network,'' \emph{eTransportation}, p. 100615, 2026.

\bibitem{liu2026attacking}
D.~Liu, X.~Cai, J.~Dong, Z.~Guo, X.~Qu, R.~Guan, X.~Fang, and D.~Ye, ``Attacking gray-box large vision-language models with adaptive svd-structured adversarial alignment,'' in \emph{International Conference on Machine Learning}, 2026.

\bibitem{ai2026attacking}
N.~Ai, G.~Chen, X.~Cai, Z.~Guo, D.~Liu, P.~Zhou, and O.~Arandjelovi{\'c}, ``Attacking hard-label large vision-language models with model-sensitive adversarial patch designs,'' \emph{Pattern Recognition}, p. 114261, 2026.

\bibitem{radford2021learning}
A.~Radford, J.~W. Kim, C.~Hallacy, A.~Ramesh, G.~Goh, S.~Agarwal, G.~Sastry, A.~Askell, P.~Mishkin, J.~Clark \emph{et~al.}, ``Learning transferable visual models from natural language supervision,'' in \emph{International conference on machine learning}.\hskip 1em plus 0.5em minus 0.4em\relax PMLR, 2021, pp. 8748--8763.

\bibitem{sun2023eva}
Q.~Sun, Y.~Fang, L.~Wu, X.~Wang, and Y.~Cao, ``Eva-clip: Improved training techniques for clip at scale,'' \emph{arXiv preprint arXiv:2303.15389}, 2023.

\bibitem{qian2024enhancing}
Y.~Qian, K.~Chen, B.~Wang, Z.~Gu, S.~Ji, W.~Wang, and Y.~Zhang, ``Enhancing transferability of adversarial examples through mixed-frequency inputs,'' \emph{IEEE Transactions on Information Forensics and Security}, vol.~19, pp. 7633--7645, 2024.

\bibitem{qian2025enhancing}
Y.~Qian, J.~Sha, B.~Wang, Z.~Gu, and Y.~Zhang, ``Enhancing transferability of targeted adversarial examples through amplitude spectrum alignment,'' \emph{Multimedia Systems}, vol.~31, no.~5, p. 345, 2025.

\bibitem{yin2019fourier}
D.~Yin, R.~Gontijo~Lopes, J.~Shlens, E.~D. Cubuk, and J.~Gilmer, ``A fourier perspective on model robustness in computer vision,'' \emph{Advances in Neural Information Processing Systems}, vol.~32, 2019.

\bibitem{wang2020high}
H.~Wang, X.~Wu, Z.~Huang, and E.~P. Xing, ``High-frequency component helps explain the generalization of convolutional neural networks,'' in \emph{Proceedings of the IEEE/CVF conference on computer vision and pattern recognition}, 2020, pp. 8684--8694.

\bibitem{liu2023point}
D.~Liu, W.~Hu, and X.~Li, ``Point cloud attacks in graph spectral domain: When 3d geometry meets graph signal processing,'' \emph{IEEE Transactions on Pattern Analysis and Machine Intelligence}, 2023.

\bibitem{hu2022exploring}
Q.~Hu, D.~Liu, and W.~Hu, ``Exploring the devil in graph spectral domain for 3d point cloud attacks,'' in \emph{European Conference on Computer Vision}.\hskip 1em plus 0.5em minus 0.4em\relax Springer, 2022, pp. 229--248.

\bibitem{luo2022frequency}
C.~Luo, Q.~Lin, W.~Xie, B.~Wu, J.~Xie, and L.~Shen, ``Frequency-driven imperceptible adversarial attack on semantic similarity,'' in \emph{Proceedings of the IEEE/CVF conference on computer vision and pattern recognition}, 2022, pp. 15\,315--15\,324.

\bibitem{geirhos2018imagenet}
R.~Geirhos, P.~Rubisch, C.~Michaelis, M.~Bethge, F.~A. Wichmann, and W.~Brendel, ``Imagenet-trained cnns are biased towards texture; increasing shape bias improves accuracy and robustness,'' in \emph{International conference on learning representations}, 2018.

\bibitem{zhao2021understanding}
D.~Zhao, A.~Wang, and O.~Russakovsky, ``Understanding and evaluating racial biases in image captioning,'' in \emph{Proceedings of the IEEE/CVF international conference on computer vision}, 2021, pp. 14\,830--14\,840.

\bibitem{cui2024robustness}
X.~Cui, A.~Aparcedo, Y.~K. Jang, and S.-N. Lim, ``On the robustness of large multimodal models against image adversarial attacks,'' in \emph{Proceedings of the IEEE/CVF Conference on Computer Vision and Pattern Recognition}, 2024, pp. 24\,625--24\,634.

\bibitem{zhou2022learning}
K.~Zhou, J.~Yang, C.~C. Loy, and Z.~Liu, ``Learning to prompt for vision-language models,'' \emph{International Journal of Computer Vision}, vol. 130, no.~9, pp. 2337--2348, 2022.

\bibitem{li2024one}
L.~Li, H.~Guan, J.~Qiu, and M.~Spratling, ``One prompt word is enough to boost adversarial robustness for pre-trained vision-language models,'' in \emph{Proceedings of the IEEE/CVF Conference on Computer Vision and Pattern Recognition}, 2024, pp. 24\,408--24\,419.

\bibitem{zhu2023minigpt}
D.~Zhu, J.~Chen, X.~Shen, X.~Li, and M.~Elhoseiny, ``Minigpt-4: Enhancing vision-language understanding with advanced large language models,'' \emph{arXiv preprint arXiv:2304.10592}, 2023.

\bibitem{wang2024qwen2}
P.~Wang, S.~Bai, S.~Tan, S.~Wang, Z.~Fan, J.~Bai, K.~Chen, X.~Liu, J.~Wang, W.~Ge \emph{et~al.}, ``Qwen2-vl: Enhancing vision-language model's perception of the world at any resolution,'' \emph{arXiv preprint arXiv:2409.12191}, 2024.

\bibitem{chen2024internvl}
Z.~Chen, J.~Wu, W.~Wang, W.~Su, G.~Chen, S.~Xing, M.~Zhong, Q.~Zhang, X.~Zhu, L.~Lu \emph{et~al.}, ``Internvl: Scaling up vision foundation models and aligning for generic visual-linguistic tasks,'' in \emph{Proceedings of the IEEE/CVF Conference on Computer Vision and Pattern Recognition}, 2024, pp. 24\,185--24\,198.

\bibitem{deng2009imagenet}
J.~Deng, W.~Dong, R.~Socher, L.-J. Li, K.~Li, and L.~Fei-Fei, ``Imagenet: A large-scale hierarchical image database,'' in \emph{2009 IEEE conference on computer vision and pattern recognition}.\hskip 1em plus 0.5em minus 0.4em\relax Ieee, 2009, pp. 248--255.

\bibitem{liang2024revisiting}
S.~Liang, J.~Liang, T.~Pang, C.~Du, A.~Liu, E.-C. Chang, and X.~Cao, ``Revisiting backdoor attacks against large vision-language models,'' \emph{arXiv preprint arXiv:2406.18844}, 2024.

\bibitem{wang2025attention}
X.~Wang, S.~Wang, Z.~Ge, Y.~Luo, and S.~Zhang, ``Attention! you vision language model could be maliciously manipulated,'' \emph{arXiv preprint arXiv:2505.19911}, 2025.

\bibitem{zhao2023svit}
B.~Zhao, B.~Wu, M.~He, and T.~Huang, ``Svit: Scaling up visual instruction tuning,'' \emph{arXiv preprint arXiv:2307.04087}, 2023.

\bibitem{ramesh2021zero}
A.~Ramesh, M.~Pavlov, G.~Goh, S.~Gray, C.~Voss, A.~Radford, M.~Chen, and I.~Sutskever, ``Zero-shot text-to-image generation,'' in \emph{International conference on machine learning}.\hskip 1em plus 0.5em minus 0.4em\relax Pmlr, 2021, pp. 8821--8831.

\bibitem{anthropic2024claude}
\BIBentryALTinterwordspacing
{Anthropic}, ``Claude 3.5 sonnet,'' June 2024. [Online]. Available: \url{https://www.anthropic.com/news/claude-3-5-sonnet}
\BIBentrySTDinterwordspacing

\bibitem{anthropic2025claude}
\BIBentryALTinterwordspacing
------, ``Claude 3.7 sonnet: Hybrid reasoning model,'' February 2025. [Online]. Available: \url{https://www.anthropic.com/news/claude-3-7-sonnet}
\BIBentrySTDinterwordspacing

\bibitem{openai2024gpt4o}
\BIBentryALTinterwordspacing
{OpenAI}, ``Gpt-4o: Real-time multimodal reasoning,'' May 2024. [Online]. Available: \url{https://openai.com/blog/hello-gpt-4o}
\BIBentrySTDinterwordspacing

\bibitem{openai2025gpt4}
\BIBentryALTinterwordspacing
------, ``Gpt-4.1: Long context and cost-efficient coding,'' April 2025. [Online]. Available: \url{https://openai.com/blog/gpt-4-1-release}
\BIBentrySTDinterwordspacing

\bibitem{deepmind2024gemini2}
\BIBentryALTinterwordspacing
{Google DeepMind}, ``Gemini 2.0: Native multimodal agents,'' December 2024. [Online]. Available: \url{https://blog.google/technology/gemini/gemini-2-0-flash}
\BIBentrySTDinterwordspacing

\bibitem{li2025frustratingly}
Z.~Li, X.~Zhao, D.-D. Wu, J.~Cui, and Z.~Shen, ``A frustratingly simple yet highly effective attack baseline: Over 90\% success rate against the strong black-box models of gpt-4.5/4o/o1,'' \emph{arXiv preprint arXiv:2503.10635}, 2025.

\bibitem{zhao2026pushing}
X.~Zhao, Z.~Li, Y.~Luo, J.~Cui, and Z.~Shen, ``Pushing the frontier of black-box lvlm attacks via fine-grained detail targeting,'' \emph{arXiv preprint arXiv:2602.17645}, 2026.

\bibitem{jia2025adversarial}
X.~Jia, S.~Gao, S.~Qin, T.~Pang, C.~Du, Y.~Huang, X.~Li, Y.~Li, B.~Li, and Y.~Liu, ``Adversarial attacks against closed-source mllms via feature optimal alignment,'' \emph{arXiv preprint arXiv:2505.21494}, 2025.

\bibitem{james1980monte}
F.~James, ``Monte carlo theory and practice,'' \emph{Reports on progress in Physics}, vol.~43, no.~9, p. 1145, 1980.

\bibitem{zhu2023enhancing}
M.~Zhu, S.~Wei, L.~Shen, Y.~Fan, and B.~Wu, ``Enhancing fine-tuning based backdoor defense with sharpness-aware minimization,'' in \emph{Proceedings of the IEEE/CVF International Conference on Computer Vision}, 2023, pp. 4466--4477.

\bibitem{shi2023black}
Y.~Shi, M.~Du, X.~Wu, Z.~Guan, J.~Sun, and N.~Liu, ``Black-box backdoor defense via zero-shot image purification,'' \emph{Advances in Neural Information Processing Systems}, vol.~36, pp. 57\,336--57\,366, 2023.

\bibitem{yan2020higcin}
R.~Yan, L.~Xie, J.~Tang, X.~Shu, and Q.~Tian, ``Higcin: Hierarchical graph-based cross inference network for group activity recognition,'' \emph{IEEE transactions on pattern analysis and machine intelligence}, vol.~45, no.~6, pp. 6955--6968, 2020.

\bibitem{yan2023progressive}
R.~Yan, L.~Xie, X.~Shu, L.~Zhang, and J.~Tang, ``Progressive instance-aware feature learning for compositional action recognition,'' \emph{IEEE Transactions on Pattern Analysis and Machine Intelligence}, vol.~45, no.~8, pp. 10\,317--10\,330, 2023.

\bibitem{qu2026spatio}
H.~Qu, X.~Shu, R.~Yan, H.~Gao, W.~Wang, and J.~Tang, ``Spatio-temporal decoupled knowledge compensator for few-shot action recognition,'' \emph{IEEE Transactions on Pattern Analysis and Machine Intelligence}, 2026.

\bibitem{xing2026vision}
L.~Xing, A.~J. Wang, R.~Yan, X.~Shu, and J.~Tang, ``Vision-centric token compression in large language model,'' \emph{Advances in Neural Information Processing Systems}, vol.~38, pp. 33\,080--33\,110, 2026.

\bibitem{xing2025locality}
L.~Xing, H.~Qu, R.~Yan, X.~Shu, and J.~Tang, ``Locality-aware cross-modal correspondence learning for dense audio-visual events detection,'' \emph{IEEE Transactions on Circuits and Systems for Video Technology}, 2025.

\bibitem{qu2025mvp}
H.~Qu, R.~Yan, X.~Shu, H.~Gao, P.~Huang, and G.-S. Xie, ``Mvp-shot: Multi-velocity progressive-alignment framework for few-shot action recognition,'' \emph{IEEE Transactions on Multimedia}, 2025.

\end{thebibliography}

\end{document}